\documentclass{article}

\usepackage{microtype}
\usepackage{graphicx}
\usepackage{subfigure}
\usepackage{sectsty}
\usepackage{latexsym}
\usepackage{amssymb}
\usepackage{amsmath}
\usepackage{amsthm}
\usepackage{booktabs} 
\usepackage{multirow}

\usepackage{hyperref}

\usepackage[accepted]{icml2023}

\usepackage{amsmath}
\usepackage{amssymb}
\usepackage{mathtools}
\usepackage{amsthm}
\usepackage{bm}

\usepackage[capitalize,noabbrev]{cleveref}

\DeclareMathOperator*{\argmax}{arg\,max}  

\usepackage[ruled, vlined, linesnumbered]{algorithm2e}
\usepackage{algpseudocode}
\algdef{SE}[SUBALG]{Indent}{EndIndent}{}{\algorithmicend\ }%
\algtext*{Indent}
\algtext*{EndIndent}
\algnewcommand\algorithmicforeach{\textbf{for each}}
\algdef{S}[FOR]{ForEach}[1]{\algorithmicforeach\ #1\ \algorithmicdo}
\SetKwComment{Comment}{$\triangleright$\ }{}

\begin{document}

\twocolumn[
\icmltitle{GraphXForm: Graph transformer for computer-aided molecular design}



\icmlsetsymbol{equal}{*}

\begin{icmlauthorlist}
\icmlauthor{Jonathan Pirnay}{equal,a,b}
\icmlauthor{Jan G. Rittig}{equal,c}
\icmlauthor{Alexander B. Wolf}{equal,d}
\icmlauthor{Martin Grohe}{e}
\icmlauthor{Jakob Burger}{d}
\icmlauthor{Alexander Mitsos}{f,c,g}
\icmlauthor{Dominik G. Grimm}{a,b}
\end{icmlauthorlist}

\icmlaffiliation{a}{Technical University of Munich, TUM Campus Straubing for Biotechnology and Sustainability, Bioinformatics, Straubing}
\icmlaffiliation{b}{University of Applied Sciences Weihenstephan-Triesdorf, Bioinformatics, Straubing}
\icmlaffiliation{c}{Process Systems Engineering (AVT.SVT), RWTH Aachen University, Aachen}
\icmlaffiliation{d}{Technical University of Munich, TUM Campus Straubing for Biotechnology and Sustainability, Laboratory of Chemical Process Engineering, Straubing}
\icmlaffiliation{e}{Chair of Computer Science 7, RWTH Aachen University, Aachen}
\icmlaffiliation{f}{JARA Center for Simulation and Data Science (CSD), Aachen}
\icmlaffiliation{g}{Forschungszentrum Jülich GmbH, Institute of Climate and Energy Systems ICE-1: Energy Systems Engineering, Jülich}

\icmlcorrespondingauthor{Dominik G. Grimm}{dominik.grimm@hswt.de}

\icmlkeywords{Machine Learning, Molecule Design}

\vskip 0.3in
]



\printAffiliationsAndNotice{\icmlEqualContribution} 
\begin{@twocolumnfalse}
\begin{abstract}
Generative deep learning has become pivotal in molecular design for drug discovery, materials science, and chemical engineering. A widely used paradigm is to pretrain neural networks on string representations of molecules and fine-tune them using reinforcement learning on specific objectives. However, string-based models face challenges in ensuring chemical validity and enforcing structural constraints like the presence of specific substructures. We propose to instead combine graph-based molecular representations, which can naturally ensure chemical validity, with transformer architectures, which are highly expressive and capable of modeling long-range dependencies between atoms. Our approach iteratively modifies a molecular graph by adding atoms and bonds, which ensures chemical validity and facilitates the incorporation of structural constraints. We present GraphXForm, a decoder-only graph transformer architecture, which is pretrained on existing compounds and then fine-tuned using a new training algorithm that combines elements of the deep cross-entropy method and self-improvement learning.
We evaluate GraphXForm on various drug design tasks, demonstrating superior objective scores compared to state-of-the-art molecular design approaches. Furthermore, we apply GraphXForm to two solvent design tasks for liquid–liquid extraction, again outperforming alternative methods while flexibly enforcing structural constraints or initiating design from existing molecular structures.
\end{abstract}
\end{@twocolumnfalse}

\section{Introduction}

Molecular design plays an important role across many fields, such as drug discovery, materials science, and chemical engineering. The immense chemical search space, estimated to contain between $10^{60}$ and $10^{100}$ potential molecules \cite{Schneider_2005}, renders manual approaches to molecular design both arduous and resource-intensive.

Advancements in deep learning have significantly impacted molecular design, enabling efficient navigation of the chemical space with the help of neural networks \cite{fu2022reinforced,Gomez-Bombarelli2016SmilesVAE,segler2018generating,Bjerrum2017MolecularGW,zhang2023artificial}. 
A prevalent paradigm is to represent molecules as strings of text, such as SMILES \cite{doi:10.1021/ci00057a005} or SELFIES \cite{krenn2020self}, and use neural network architectures from language modeling, such as recurrent neural networks (RNNs) or transformers \cite{vaswani2017attention}, to generate novel molecular structures. These architectures are typically pretrained on large datasets of existing molecules to learn general underlying patterns in the strings and then fine-tuned on specific objective functions via reinforcement learning (RL) for downstream tasks \cite{olivecrona2017molecular,gao2022sample,xu2024reinventtransformer,mazuz2023molecule,gupta2018generative}. 

To address the challenge of capturing long-range dependencies in sequential data that RNNs face, transformers \cite{vaswani2017attention} have been successfully applied due to their ability to model long-range dependencies more efficiently and as they are widely used in language modeling today \cite{xu2024reinventtransformer}. However, transformers are resource-intensive,  and fine-tuning them with RL further constrains the model sizes that are practical in real-world applications \cite{xu2024reinventtransformer,henderson2018deep,schulman2017proximal}. In general, chemical language models may propose string representations of molecules with invalid chemical structures – for example, when SMILES syntax or valence constraints are violated – which has led to numerous works aimed at circumventing this problem \cite{o2018deepsmiles,krenn2020self,cheng2023group,dai2018syntaxdirected}. Despite recent evidence that invalid SMILES can actually be beneficial from a language modeling perspective \cite{skinnider2024invalid}, they can harm the RL component of the pipeline as they can increase sample complexity and necessitate reward shaping to account for them. Moreover, the sequential nature of string synthesis makes it challenging to enforce structural constraints – such as ensuring a minimum number of specific atom types, restricting bonding patterns, or initiating design from predefined molecular substructures – often requiring scaffold-constrained techniques \cite{doi:10.1021/acs.jcim.0c01015,arus2020smiles}.

An alternative approach is to represent molecules directly as graphs, where atoms are nodes and bonds are edges, and to develop models that modify a molecular graph or design it directly \cite{Wang_2023,li2018multi,Jin2018,zhou2019optimization,jensen2019graph,de2018molgan,zang2020moflow,Zhang_2023,mahmood2021masked,maziarka2019molecule, rittig2023graph}. Working at the graph level allows explicit encoding of atomic interactions and bonding rules, ensuring chemical validity, and makes it straightforward to start from existing structures and modify them. For example, graph-based methods like GraphGA \cite{jensen2019graph} employ genetic algorithms to modify molecular graphs directly, even outperforming several neural-based string-synthesis methods without relying on neural networks. As a deep learning example, \citet{zhou2019optimization} use deep Q-Learning \cite{mnih2013playing} with a simple feedforward network to optimize graphs from scratch by adding or removing atoms and bonds.

We propose to combine and extend the strengths of both paradigms: employing transformers for their ability to capture long-range dependencies in sequence data, and leveraging pretraining on existing molecules, all while working directly on the molecular graph. More specifically, we use graph transformers \cite{ying2021transformers,maziarka2019molecule} and formulate molecule design as a sequential task, where a molecular graph – starting from an arbitrary structure – is iteratively extended by placing atoms and adding bonds. 

To this end, we introduce GraphXForm, a decoder-only graph transformer architecture that guides these incremental decisions, predicting the next modification based on the current molecular graph. Pretrained on existing compounds, we propose a fine-tuning approach for downstream tasks that combines elements of the deep cross-entropy method \cite{wagner2021constructions} and self-improvement learning \cite{pirnay2024take}. Unlike commonly used deep RL methods like REINFORCE \cite{williams1992simple}, this approach allows for stable training of deep transformers with many layers. 

We test GraphXForm for two molecular applications: (1) drug development and (2) solvent design.
While algorithmic advances in molecular design have been primarily focused on \textit{de novo} drug development and corresponding benchmarks \cite{Brown_2019, polykovskiy2020molecular, gao2022sample, thomas2024molscore}, other areas such as materials science and chemical engineering have recently gained attention.
Recent applications include catalyst \cite{schilter2023designing}, fuel \cite{rittig2023graph,sarathy2024artificial}, polymer \cite{jiang2024property,yue2024benchmarking}, surfactant \cite{nnadili2024surfactant}, chemical reaction substrate \cite{nigam2023tartarus}, and solvent \cite{konig2024machine} design.
Thus, we test GraphXForm in both established and newer application areas.

First, we consider drug design by evaluating GraphXForm on the goal-directed tasks of the well-established GuacaMol benchmark \cite{Brown_2019}.
The benchmark includes multiple design tasks such as drug rediscovery, isomer identification, and multi-property optimization.
Based on these different design objectives, we demonstrate the soundness of GraphXForm and its competitive performance with state-of-the-art molecular design approaches.

Secondly, we apply GraphXForm for the design of solvents, which play a vital role in industrial chemical processes such as reactions, separations, and extractions.
While K{\"o}nig-Mattern \cite{konig2024machine} recently applied a graph-based genetic algorithm for solvent design, the use of generative ML-based approaches remains underexplored. 
Therefore, our goal is to compare newer design methods and expand their capabilities by considering molecular structure constraints.
Specifically, we evaluate GraphXForm on two liquid-liquid extraction tasks. 
The objective function in these tasks is defined by a separation factor based on activity coefficients at infinite dilution. 
For both downstream tasks, we compare GraphXForm to state-of-the-art molecular design approaches (Graph~GA \cite{jensen2019graph}, REINVENT-Transformer \cite{xu2024reinventtransformer}, Junction Tree VAE \cite{Jin2018}, and STONED \cite{nigam2021beyond}). 
Additionally, we demonstrate GraphXForm's flexibility and stability by incorporating structural constraints conceptually suited for solvent design, such as preventing certain bonding patterns or preserving molecular substructures, allowing GraphXForm to propose design candidates with highly tailored properties. 
This capability highlights the strength of our approach in tackling design tasks that are difficult for existing methods.

Our contributions are summarized as follows:
\begin{itemize}
    \item {\bf We formulate molecular design as a sequential task}, where an initial structure (e.g., a single atom) is iteratively modified by adding atoms and bonds.
    \item {\bf We introduce a graph transformer architecture} that takes a molecular graph as input and outputs probability distributions for atom and bond placement. This approach maintains the notion of using transformers for molecular design while moving away from string-based methods.
    \item {\bf We propose a training algorithm} that enables the stable and efficient fine-tuning of deep graph transformers on downstream tasks.
    \item {\bf We demonstrate that our method outperforms state-of-the-art molecular design techniques} on well-established drug design benchmarks and two solvent design tasks.
    \item {\bf We show that our method can be easily adapted to meet structural constraints} by preserving or excluding specific molecular moieties and starting the design from initial structures.
\end{itemize}

Our code for pretraining and fine-tuning is available at: \\\url{https://github.com/grimmlab/graphxform}.

\section{Methods and Modeling}

In this section, we introduce GraphXForm. In Section~\ref{sec:problem_setup}, we formally set up the molecular design task as the sequential construction of a graph. As in a deep reinforcement learning setup, the goal is to find a policy network that guides these sequential decisions. In Section~\ref{sec:learning_algo}, we introduce our algorithm for training the policy network, before describing the used transformer architecture in Section~\ref{sec:network}.

\subsection{Molecular design}\label{sec:problem_setup}

\paragraph*{Molecular graph} In the following, we represent a molecule by its hydrogen-suppressed graph representation, where nodes correspond to atoms and edges correspond to bonds. For ease of notation, we assume an arbitrary ordering over the nodes. Let $\Sigma = (\Sigma_1, \dots, \Sigma_k)$ be an underlying alphabet of possible atom types. We represent a molecule with $n$ atoms by a pair $(\bm a, \bm B)$, where $\bm a = (a_1, \dots, a_n) \in \{1, \dots, k\}^n$ and $a_i$ indicates that the $i$-th node is of atom type $\Sigma_{a_i}$. The matrix $\bm B = (B_{ij})_{1 \leq i,j \leq n} \in \mathbb N_0^{n \times n}$ represents the bonds and their orders between atoms, i.e., we have that $B_{ij} \in \mathbb N_0$ is the bond order between the $i$-th and the $j$-th atom. In particular, $\bm B$ is symmetric with zero diagonal and nonzero columns and rows (i.e., each atom is connected to at least one other atom and not to itself). For example, given the alphabet $\Sigma = (\mathrm C, \mathrm N, \mathrm O)$ the molecule with SMILES representation C=O can be given as $(\bm a, \bm B)$ with $\bm a = (1, 3)$ and $\bm B = \begin{pmatrix} 0 & 2 \\ 2 & 0 \end{pmatrix}$. 

We denote by $\mathcal{M'}$ the space of all molecular graphs $m=(\bm a, \bm B)$ as described above. Accordingly, let $\mathcal M \subset \mathcal{M'}$ be the subspace of molecular graphs that are chemically valid, that is, all non-hydrogen atoms follow the octet rule. We refer to following the octet rule as satisfying the 'valence constraints' throughout the paper. 
In this framework, ionization can be incorporated straightforwardly by extending the alphabet with additional atom types that allow for an adjusted number of non-hydrogen bonds. For instance, to model ionized carbon atoms, we can include the types $\mathrm{C^+}$ and $\mathrm{C^-}$ in $\Sigma$, which are designated with a positive or negative charge and are permitted to form up to five and three non-hydrogen bonds, respectively. Chirality can be handled in an analogous manner. We refer to the Supplementary Information for the complete alphabet used.

\paragraph*{Sequential molecular graph design} Similar to \citet{zhou2019optimization}, we pose the molecular design as a sequential Markov decision process (MDP), where an agent assembles a molecular graph by iteratively adding atoms or bonds between atoms. We note that in the graph, hydrogens are always only considered implicitly, and an addition of an atom or a bond leads to a replacement of implicit hydrogen. 

On a high level, a single molecule is constructed by the agent as follows: The agent observes an initial molecule $m^{(0)}$ and then performs some {\em action} $x^{(0)}$ on it, resulting in a new molecule $m^{(1)}$. Then, the agent observes $m^{(1)}$, decides on an action $x^{(1)}$ leading to molecule $m^{(2)}$, and so on. This iterative design process ends on some molecule $m^{(T)}$ once the agent decides to perform a special action $\textsc{DontChange}$, which does not alter the molecular graph but rather marks the design as {\em completed}. 

We formalize this as follows.
Let $m^{(t)} = (\bm a^{(t)}, \bm B^{(t)})$ be some molecule with atoms $\bm a^{(t)} \in \{1,\dots,k\}^{n^{(t)}}$ and bond matrix $\bm B^{(t)} \in \mathbb N_0^{n^{(t)} \times n^{(t)}}$ as above. We transition to a new molecule $m^{(t+1)} = (\bm a^{(t+1)}, \bm B^{(t+1)})$ by performing some {\em action} $x^{(t)} \in \mathcal A$ on $m^{(t)}$. An action $x^{(t)}$ in the action space $\mathcal A$ is of one of the following three types:
\begin{enumerate}
    \item $x^{(t)} = \textsc{DontChange}$, which terminates the design of the molecule. In particular, $m^{(t+1)} = m^{(t)}$ and $n^{(t+1)} = n^{(t)}$. 
    \item $x^{(t)} = \textsc{AddAtom}(j,l,o)$, with $j \in \{1, \dots, k\}$, $l \in \{1, \dots, n^{(t)}\}$, and $o \in \mathbb N$. This action adds an atom of type $\Sigma_j$ to the graph and connects it to the $l$-th atom with a bond of order $o$. In particular, we set $n^{(t+1)} = n^{(t)} + 1$. For the new molecule $m^{(t+1)} = (\bm a^{(t+1)}, \bm B^{(t+1)})$, the atom vector $\bm a^{(t+1)} \in \mathbb N^{n^{(t+1)}}$ is obtained by appending $j$ to $\bm a^{(t)}$. The second entry $l \in \{1, \dots, n^{(t)}\}$ indicates that we bond this new $(n+1)$-th atom to the $l$-th atom with order $o$, i.e., $\bm B^{(t+1)} \in \mathbb N_0^{n^{(t+1)} \times n^{(t+1)}}$ is obtained by appending a zero row and column to $\bm B^{(t)}$ and setting $B^{(t+1)}_{n+1, l} = B^{(t+1)}_{l, n+1} = o$.
    \item $x^{(t)} = \textsc{AddBond}(j, l, o)$, meaning that we are adding a bond of order $o \in \mathbb N$ between {\em existing, unbonded} atoms $j$ and $l$. In particular, we have $n^{(t+1)} = n^{(t)}$ and $\bm a^{(t+1)} = \bm a^{(t)}$. The bond matrix $\bm B^{(t+1)}$ is obtained from $\bm B^{(t)}$ by setting $B^{(t+1)}_{j, l} = B^{(t+1)}_{l, j} = o$. 
\end{enumerate}

Given a molecule $m^{(0)}$ and a sequence of actions $x^{(0)}, \dots, x^{(t-1)}$, we will also write $(m^{(0)}, x^{(0)}, \dots, x^{(t-1)})$ for the molecule $m^{(t)} \in \mathcal M$ that results from $m^{(0)}$ by sequentially applying the actions $x^{(0)}, \dots, x^{(t-1)}$ to $m^{(0)}$. In general, the action sequence $x^{(0)}, \dots, x^{(t-1)}$ is not unique to get from $m^{(0)}$ to $m^{(t)}$.

\begin{figure*}
    \center
  \includegraphics[width=0.72\linewidth]{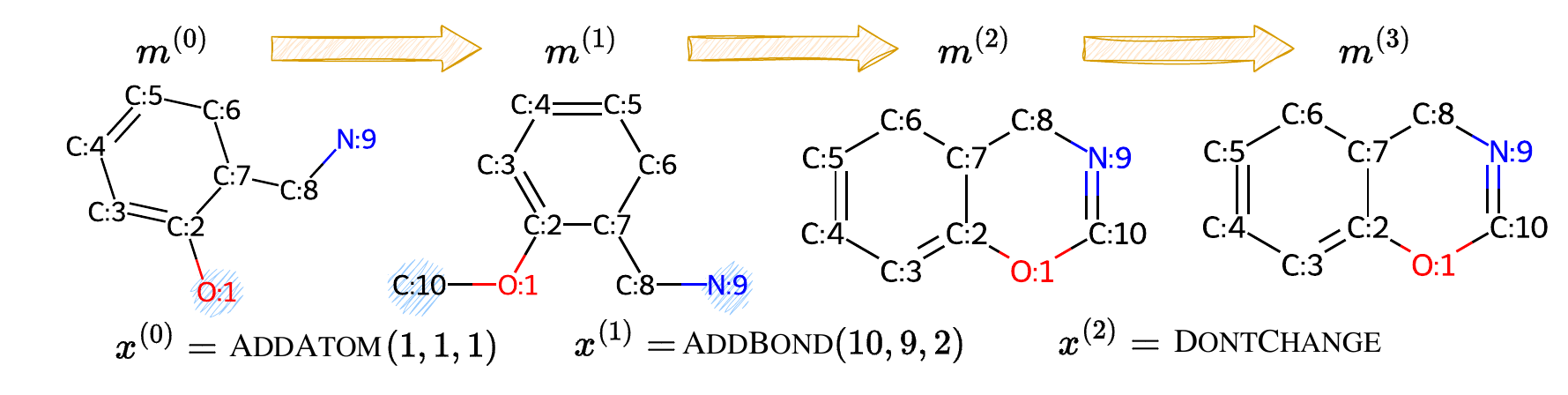}
  \caption{Example for the sequential application of actions $x^{(0)}, x^{(1)}, x^{(2)}$ to a molecule $m^{(0)}$, using the alphabet $\Sigma = (\mathrm C, \mathrm N, \mathrm O)$. We show the index of each atom, which can be arbitrarily chosen at the beginning. Light blue indicates where in the graph an action is applied. The last action is $\textrm{\textsc{DontChange}}$, which does not change the molecular graph, but marks it as a complete design.}
  \label{fig:stepwise_construction}
\end{figure*}

Starting from a molecule in $\mathcal M$, by removing chemically invalid actions from the action space (which would lead to violation of valence constraints) at each step, we can guarantee staying in the space of chemically valid molecules. Once the agent chooses the action $\textsc{DontChange}$, the design process is considered complete. 
We note that starting from an appropriate initial atom, it is in fact possible to reach every target molecule in the chemically valid space $\mathcal M$ (i.e., all molecular graphs that can be constructed from the alphabet $\Sigma$ and that satisfy the valence constraints) when starting from any atom that exists in the target molecule.
For an illustrative molecule construction, see Fig.~\ref{fig:stepwise_construction}.

\paragraph*{Molecular optimization}
Formally, we aim to design molecules
\begin{equation}\label{eq:mol_optimization}
    m^* \in \argmax_{m \in \mathcal M} f(m)
\end{equation}
that maximize a predefined objective function $f \colon \mathcal M' \to \mathbb R \cup \{-\infty\}$, where chemically invalid molecules are mapped to $-\infty$. As in previous work \cite{mazuz2023molecule,zhou2019optimization,xu2024reinventtransformer}, we pose the corresponding learning problem to (\ref{eq:mol_optimization}) in the terms of deep RL: the agent's decision at each step is guided by a {\em policy} that maps a chemically valid molecule to a probability distribution over possible actions. The policy is modeled by a neural network: We write $\pi_\theta \colon \mathcal M \to \Delta \mathcal A$ for a policy depending on network parameters $\theta$, that maps a valid molecule $m \in \mathcal M$ to a probability distribution $\pi_\theta(m)$ over $\mathcal A$. 
The goal is to find $\pi_\theta$ that, given any initial molecule $m^{(0)}$, maximizes the expectation
\begin{equation}
    \mathbb E_{\substack{x^{(0)}, \dots, x^{(T)} \\ x^{(T)} = \textsc{DontChange}} \sim \pi_\theta} \left[ f\left((m^{(0)}, x^{(0)}, \dots, x^{(T)})\right) \right],
\end{equation}
where the expectation is taken over finished molecules sampled from $\pi_{\theta}$. 

\paragraph*{Satisfying constraints} To ensure chemical validity at every step of the molecule design process, we mask any action in the policy that would lead to a violation of valence constraints. That means, that we set the corresponding probability to zero in the distribution predicted by the policy (before re-normalizing the distribution).

Our graph-based approach allows us to extend this concept of action masking to enforce additional constraints, such as particular bonding patterns, minimum/maximum number of atoms and their types, or restricting structural motifs like rings. We explore these constraints in detail in Section~\ref{sec:experiments}. We note that it is possible to simply assign an objective value of $- \infty$ to molecules that violate these constraints {\em after} the design process. However, the ability to preemptively avoid constraint-violating regions by masking actions {\em during} the design process reduces the search space.

\subsection{Learning algorithm}\label{sec:learning_algo}

We now introduce our method for training the policy neural network $\pi_\theta$. Our proposed learning algorithm is a hybrid of the deep cross-entropy method (CEM) \cite{wagner2021constructions} and self-improvement learning (SIL) \cite{huang2023large,corsini2024self,pirnay2024selfimprovement,pirnay2024take}, a sampling-based approach to expert iteration \cite{anthony2017thinking}. Both the deep CEM and SIL train the neural network over multiple epochs in a self-improving loop. In each epoch, the current policy is used to generate a set of action sequences, from which the best sequences are selected as 'pseudo-labels' to serve as the dataset for {\em supervised} training. The network is then trained for one epoch to assign higher probabilities to these sequences, and the updated network is subsequently used to generate new action sequences. Unlike many deep RL algorithms, this approach does not require reward shaping or value approximation. Moreover, training in a supervised way provides stability and facilitates the use of larger, decoder-only transformer architectures \cite{vaswani2017attention}. 

There are key differences between SIL and the deep CEM. The deep CEM is formulated for problems with a {\em single} instance (as in our molecular design task) and retains a fixed percentage of the best action sequences in each epoch, which are obtained through simple sampling from the policy's predicted distributions at each step. In contrast, SIL is typically applied to problems with {\em infinitely} many instances and employs more advanced sequence decoding techniques – such as sequence sampling without replacement \cite{kool2019stochastic} – to improve solution quality and diversity while maintaining efficient decoding speed.

In our proposed approach, we seek to develop a method that works for single-instance problems, as in the deep CEM, while retaining the diverse sampling mechanism used in SIL methods \cite{pirnay2024selfimprovement,pirnay2024take}. In particular, we adopt the {\em Take a Step and Reconsider} (TASAR) method \cite{pirnay2024take} to sample action sequences, as detailed below. The pseudocode for our approach is presented in Algorithm~\ref{algo:learning}, and we describe the key steps as follows:

\begin{enumerate}
    \item We begin with a policy network $\pi_\theta$. The parameters $\theta$ can be initialized randomly or, e.g., pretrained in an unsupervised manner on existing molecules (see Section~\ref{sec:training_setup}). Additionally, we assume an initial molecule $m_0 \in \mathcal M$ in the space of chemically valid molecules $\mathcal M$, from which the construction starts. In practice, we typically choose $m_0$ to consist of a single carbon atom.
    \item Throughout training, we maintain a set $\textsc{BestFound}$ containing the best molecules discovered so far. This set is initially initialized with the starting molecule $m_0$.
    \item In each epoch, we sample action sequences from the policy using the TASAR method \cite{pirnay2024take}. Specifically, $\beta$ action sequences are sampled {\em without replacement} using Stochastic Beam Search \cite{kool2019stochastic} with a beam width $\beta \in \mathbb N$. These sequences are then evaluated using the objective function. The best action sequence among the $\beta$ samples is followed for a predefined number $\sigma$ of actions. Subsequently, alternative, previously {\em unseen} actions are sampled from the resulting partial sequence to further explore and potentially improve the solution, and the process continues. Sampling without replacement ensures that unique action sequences are generated, effectively exploring the search space (particularly for shorter sequences \cite{shi2020incremental}). Although different action sequences may result in the same molecule, this does not pose an issue for TASAR since the policy is concerned with generating action sequences rather than the molecular structure itself. By sampling sequences without replacement, the policy is encouraged to produce diverse outputs, even if the resulting molecules are identical.
    \item After sampling, the set $\textsc{BestFound}$ serves as the training dataset for supervised learning (lines 6-8). Similarly to how decoder-only models in language modeling are trained to predict the next token from partial text, we sample batches of intermediate molecules and train the network with a cross-entropy loss to predict the next action for the corresponding molecule.
    \item In the next epoch, the process is repeated with the updated network weights.
\end{enumerate}

\begin{algorithm}[t]
   \caption{Learning algorithm for molecular design}
   \label{algo:learning}
      \DontPrintSemicolon
      \KwIn{$\pi_\theta$: Policy network with trainable parameters $\theta$}
      \KwIn{$f$: objective function to maximize}
      \KwIn{$m_0$: initial molecule}
      \KwIn{$s \in \mathbb N$: number of best molecules to keep}
      \KwIn{$\beta, \sigma \in \mathbb N$: beam width and step size (hyperparams for TASAR \cite{pirnay2024take})}
      \BlankLine
      $\textsc{BestFound} \gets \{m_0\}$\;
      $\textsc{BestObj} \gets f(m_0)$\;
      \ForEach{\upshape epoch}{
        $\textsc{Sampled} \gets $ sample molecules with TASAR with beam width $\beta$ and step size $\sigma$\;
        $\textsc{BestFound} \gets$ top $s$ molecules in $\textsc{BestFound} \cup \textsc{Sampled}$\;
        \ForEach{\upshape batch}{
            uniformly sample from $\textsc{BestFound}$ batch of $B$ (intermediate) molecules $m_{t_{j+1}}^{(j)} = (m_{t_{j}}^{(j)}, x_{t_{j}}^{(j)})$ for $j \in \{1, \dots, B\}$\;
            minimize batch-wise cross-entropy loss $L_\theta = -\frac{1}{B}\sum_{j=1}^B \log \pi_\theta\left(x_{t_{j}}^{(j)}\ |\ m_{t_{j}}^{(j)}\right)$\;
        }
      }
      \textbf{return} $\textsc{BestFound}$
\end{algorithm}

\subsection{Policy network architecture}\label{sec:network}
The policy network receives a molecule as input and predicts a probability distribution over possible next actions using a simplified version of the Graphormer \cite{ying2021transformers}. This model treats the molecule's atoms as an unordered set of nodes and processes them through a stack of transformer layers with the attention mechanism augmented by bonding information. The resulting latent representations of the nodes are then used to predict action distributions. 

\paragraph*{Splitting actions} To make the network's predictions more fine-grained, we decompose each action (except $\textsc{DontChange}$ into three sub-actions, each determined by a separate forward pass through the network:
\begin{itemize}
    \item \textbf{Action level 0}: The agent decides whether to end the design process (choosing $\textsc{DontChange}$) or to modify the molecule. In the latter case, it selects the first atom $j$, which can be an already present one for $\textsc{AddBond}(j, l, o)$ or a new one from the alphabet for $\textsc{AddAtom}(j, l, o)$.
    \item \textbf{Action level 1}: Given a modification is intended, the agent selects a second atom $l$ from the current molecule, indicating a new bond between $j$ and $l$.
    \item \textbf{Action level 2}: Finally, the agent determines the  order $o$ of the bond between $j$ and $l$.
\end{itemize}

Figure~\ref{fig:network_architecture} provides an illustration, and we elaborate on the architecture below:

\begin{figure*}
    \center
  \includegraphics[width=\linewidth]{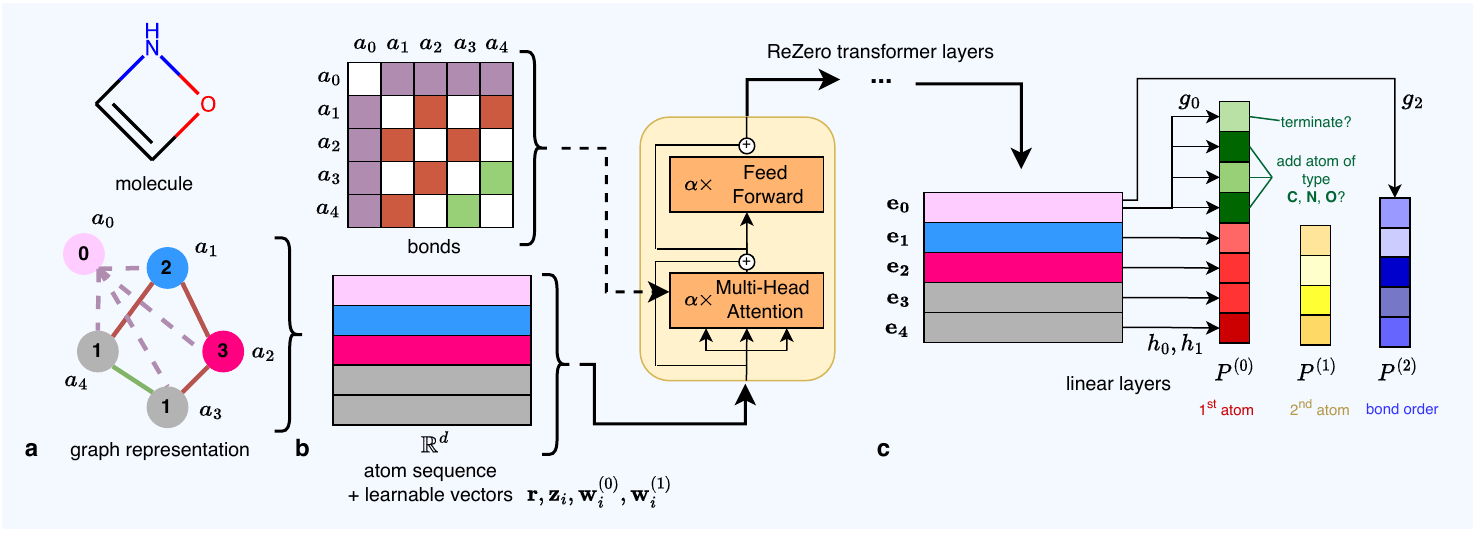}
  \caption{Flow of a molecule through the policy network of our method GraphXForm. {\bf a.} We consider the alphabet $\Sigma = (\mathrm C, \mathrm N, \mathrm O)$. The molecule's underlying graph is augmented with a virtual node (indexed with 0) and embedded into the latent space $\mathbb R^d$. Learnable vectors are added to these embeddings to encode the current action level and decisions on previous levels. {\bf b.} The latent sequence of atoms is passed through a stack of ReZero transformer layers, omitting positional encoding. In the multi-head attention, individual attention scores between atoms are biased with learnable scalars that depend on their bond order. These bias terms are learnable for each transformer layer and attention head individually. {\bf c.} The sequence output by the transformer is projected through linear layers to generate logits for the distributions $P^{(0)}, P^{(1)}$ and $P^{(2)}$.}
  \label{fig:network_architecture}
\end{figure*}

\paragraph*{Molecular graph transformer} Let $m = (\bm a, \bm B)$ represent a molecule, where $\bm a = (a_1, \dots, a_n) \in \mathbb N^n$ are the atoms, and $\bm B \in \mathbb N_0^{n \times n}$ is the bond matrix. Each atom $a_i \in \{1, \dots, k\}$ corresponds to an atom type selected from an alphabet $\Sigma = (\Sigma_1, \dots, \Sigma_k)$. 

Following the Graphormer \cite{ying2021transformers} model, we introduce a `virtual atom' $a_0 = 0$ into the molecular graph, which is connected to every atom via a special `virtual bond' (see Figure~\ref{fig:network_architecture}). This virtual bond is represented in the bond matrix by an integer outside the range of standard bond orders. The virtual atom functions similarly to the special \texttt{[CLS]}-token in BERT \cite{devlin2018bert} and accumulates sequence-level information for downstream tasks. Conceptually, the virtual atom acts as an additional message proxy between nodes in the molecular graph.

The atom sequence $(a_0, a_1, \dots, a_n)$ is first embedded latent representations $(\bm{\hat a}_0, \bm{\hat a}_1, \dots, \bm{\hat a}_n)$ in $\mathbb R^d$, where each $\bm{\hat a}_i$ is a learnable vector corresponding to the atom type. We then augment these representations with additional learnable embeddings that encode the current action level and other relevant information.

Specifically, we introduce a learnable vector $\bm{r} \in \mathbb{R}^d$ to represent the current action level. For each atom $i \in {1, \dots, n}$, we further include the embeddings $\bm{w}_i^{(0)}$ and $\bm{w}_i^{(1)}$, which indicate whether the $i$-th atom has been selected at action levels 0 and 1, respectively. Moreover, let $\bm{z}_i \in \mathbb{R}^d$ be an embedding that reflects the total number of bonds (i.e., the degree) formed by the $i$-th atom. The final augmented sequence \begin{equation}
    (\bm{\hat a}_0 + \bm r, \bm{\hat a}_1 + \bm{z}_1 + \bm{w}_1^{(0)} + \bm{w}_1^{(1)}, \dots, \bm{\hat a}_n + \bm{z}_n + \bm{w}_n^{(0)} + \bm{w}_n^{(1)})
\end{equation} is then passed through a stack of transformer layers using ReZero normalization \cite{bachlechner2021rezero}. Importantly, to maintain permutation equivariance — ensuring that the order of the atoms does not affect the outcome — we do not apply any positional encodings.

To incorporate bonding information within the transformer layers, let $(\bm A_{ij})_{0 \leq i,j \leq n} \in \mathbb R^{(n+1) \times (n+1)}$ be the computed self-attention matrix in a layer (for any attention head) corresponding to the input sequence $(a_0, a_1, \dots, a_n)$. As in Graphormer \cite{ying2021transformers}, and similar to the Molecule Attention Transformer \cite{maziarka2019molecule}, we augment the attention matrix $\bm A$ by introducing bond-specific information. This is done by adding a learnable bias to the attention scores before applying the softmax operation. Specifically, for $0 \leq i,j \leq n$, the attention score $\bm A_{ij}$ between the $i$-th and $j$-th sequence elements is modified as follows:
\begin{equation}
    \bm A_{ij} \gets \bm A_{ij} + b_{ij},
\end{equation}
where $b_{ij} \in \mathbb R$ is a learnable scalar that only depends on the bond order $B_{ij}$ for $i, j > 0$ (and not on the indices $i,j$). In particular, for the special bonds involving the virtual atom (i.e., if $i=0$ or $j=0$), $b_{ij} = b$ is a learnable scalar shared across all atoms. We note that $b_{ij}$ for each bond order is learned independently across layers and attention heads. 

\paragraph*{Action level distributions} The stack of transformer layers outputs a sequence of node embeddings $(\bm e_0, \bm e_1, \dots, \bm e_n)$, where each $\bm e_i \in \mathbb R^d$ corresponds to the original atom $a_i$. These embeddings are then used to simultaneously predict three probability distributions, $P^{(0)}$, $P^{(1)}$, and $P^{(2)}$, each corresponding to one of the action levels. The agent selects an action based on the distribution associated with the current action level.

The unnormalized log-probabilities (logits) for these distributions are computed as follows:
\begin{itemize}
    \item Action level 0 ($P^{(0)}$): 
    We denote the logits for $P^{(0)}$ as 
    \begin{equation}
        (p^{(0)}_0, p^{(0)}_1, \dots, p^{(0)}_k, q^{(0)}_{1}, \dots, q^{(0)}_{n}) \in \mathbb R^{k+1+n}.
    \end{equation}
    The first $k+1$ logits, $(p^{(0)}_0, p^{(0)}_1, \dots, p^{(0)}_k)$, are obtained by projecting the vector $\bm e_0$ through a linear layer $g_0 \colon \mathbb R^d \to \mathbb R^{k + 1}$. Here, $p^{(0)}_0$ corresponds to the $\textsc{DontChange}$ action, and for $1 \leq j \leq k$, the logit $p_j^{(0)}$ corresponds to adding a new atom of type $\Sigma_j$ (i.e., serving as the first parameter $j$ in $\textsc{AddAtom}(j,\cdot, \cdot)$. 
    The remaining logits, $q^{(0)}_{1}, \dots, q^{(0)}_n$, are computed by applying a linear layer $h_0 \colon \mathbb R^d \to \mathbb R$ independently to each of the embeddings $\bm e_1, \dots, \bm e_n$. For $1 \leq j \leq n$, the logit $q^{(0)}_{j}$ corresponds to selecting the first parameter $j$ for the $\textsc{AddBond}(j, \cdot, \cdot)$ action.
    \item Action level 1 ($P^{(1)}$): The logits $(q^{(1)}_1, \dots, q^{(1)}_n)$ for this level are obtained by applying a linear layer $h_1 \colon \mathbb R^{d} \to \mathbb R$ independently to each of $\bm e_1, \dots, \bm e_n$. Here, $q^{(1)}_l$ represents the choice of the second parameter $l$ for either $\textsc{AddAtom}(j,l,\cdot)$ or $\textsc{AddBond}(j,l,\cdot)$.
    \item Action level 2 ($P^{(2)}$): The logits $(p^{(2)}_1, \dots, p^{(2)}_y)$, which correspond to the bond order $o$ in either $\textsc{AddAtom}(j,l,o)$ or $\textsc{AddBond}(j,l,o)$, are computed by projecting $\bm e_0$ using a linear layer $g_2 \colon \mathbb R^{d} \to \mathbb R^y$, where $y$ is a predefined maximum bond order. 
\end{itemize}

After the agent selects an action at the current action level, the chosen information is fed back into the network by updating the learnable vectors $\bm r, \bm w_i^{(0)}$ and $\bm w_i^{(1)}$. This updated state is then used for the subsequent forward pass when predicting the next action level.

The multi-step action prediction allows us to easily mask actions (i.e., setting their probability to zero in the policy) that would lead to invalid molecules. While checking for actions that would violate constraints adds a small amount of computational overhead, it has a significant benefit: invalid molecules can be immediately disregarded, preventing the network from wasting resources on infeasible designs. Masking invalid actions not only reduces the search space but also speeds up training by avoiding the need for the model to learn through trial and error how to avoid invalid molecules.

\section{Results and Discussion}\label{sec:experiments}

We begin by outlining the experimental setup and hyperparameters for our method. Our first case study tackles the goal-directed design tasks from the well-established GuacaMol benchmark \cite{Brown_2019}. In our second case study, we explore two solvent design tasks by detailing the goals, their underlying objective functions, and the property prediction methods used. To contextualize our results, we also present preliminary baselines obtained by screening available molecules and compare our findings with those of other approaches.

\subsection{GraphXForm: Pretraining and fine-tuning}\label{sec:training_setup}

\paragraph*{Network hyperparameters} For all case studies and experiments, we set the latent space dimension to $d=512$ (see Section~\ref{sec:network}). The network comprises ten transformer layers with ReZero normalization, each featuring eight attention heads and a feed-forward dimension of 2048.

\paragraph*{Alphabet and pretraining} We define the atom alphabet $\Sigma$ to include C, N, O, F, P, S, Cl, Br, and I, along with variants for ionization and chirality (see Supplementary Information for the complete alphabet). Although the agent is capable of learning without prior knowledge, we pretrain the network on known molecules using SMILES strings from the ChEMBL database \cite{davies2015chembl}. We filtered the database to include only molecules containing atoms from $\Sigma$ and split it into a training set of approximately 1.5 million molecules and a validation dataset of around 70,000 molecules. Each SMILES string is converted into an action sequence in our graph formulation; since the conversion is not unique, we select one possible sequence arbitrarily. These sequences are used to train the model in a self-supervised manner – predicting the next action given previous actions – as outlined in lines 6–8 of Algorithm~\ref{algo:learning}. Training is performed with a dropout rate of 0.1, a batch size of 512, and over a total of 1.5 million batches. 

Petraining the network on existing molecules establishes a prior that captures the characteristics of real molecules, which is crucial for most generative methods such as REINVENT \cite{olivecrona2017molecular,xu2024reinventtransformer} or JT-VAE \cite{Jin2018}. 

\paragraph*{Fine-tuning} Given an objective function $f \colon \mathcal M' \to \mathbb R \cup \{-\infty\}$, we fine-tune the pretrained policy network using the learning algorithm described in Section~\ref{sec:learning_algo}. We train only the weights of the final linear layers $g_0, g_2, h_0, h_1$. Unless stated otherwise, we initialize the molecule $m_0$ as a single carbon atom. We set $s=100$ as the number of top molecules to retain throughout training. For the TASAR sampling procedure, we use a beam width of $\beta=512$ and a step size of $\sigma=12$; that is, after every four added atoms or bonds, TASAR seeks better alternative solutions. At the end of each epoch, we train the network on 20 batches of size 64, sampled uniformly from the top 100 molecules. We intentionally keep the number of training batches per epoch relatively low to prevent premature overfitting, though in most runs increasing the number of batches does not harm performance and may even speed up convergence. 

\paragraph*{Runtime and computational budget} The runtime of a method can vary significantly due to differences in implementation and available hardware. A common approach to align computational resources is to limit the number of calls to the objective function \cite{gao2022sample}. While this is useful for comparing sample efficiency, it offers limited insight into overall efficiency when objective evaluations are inexpensive. In our case studies, objective function evaluations are relatively cheap: For the solvent design tasks, for example, these are computed by evaluating the designed molecule with a surrogate neural network that can process batches in parallel. Therefore, to provide a practical comparison between our method and competing approaches, each design run is executed until convergence or until a maximum wall-clock time of eight hours is reached. We run all experiments with a single NVIDIA H100 GPU with 80 GB of memory. 

\subsection{Case study 1: Drug design}\label{subsec:design_task_1}
\paragraph*{Objectives}
We evaluate GraphXForm for {\em de novo} molecular design using the 20 goal-directed tasks from the GuacaMol benchmark \cite{Brown_2019}. 
These tasks span drug rediscovery, similarity-based design, multi-property optimization (MPO), and scaffold hopping.
We selected GuacaMol as an established benchmark because our primary focus in this study is to demonstrate the overall optimization capability of our method – i.e., achieving molecules with high scores. 
We note, however, that current comprehensive benchmarking also considers factors such as sample efficiency and molecular diversity \cite{gao2022sample,thomas2022re,thomas2024molscore}.
Such aspects could be considered in future applications and extensions of GraphXForm, e.g., by further investigating replay buffers \cite{guo2024augmented,guo2024saturn} and the TASAR parameters.

\paragraph*{Benchmark methods} 
For the drug design tasks, we compare GraphXForm against Graph~GA \cite{jensen2019graph} and REINVENT-Transformer \cite{xu2024reinventtransformer}.

Graph~GA \cite{jensen2019graph} employs a genetic algorithm that directly operates on the molecular graph, mutating atoms and fragments using crossover rules derived from graph matching. This non-learning method is highly effective at making fine-grained local modifications, and it has been shown to outperform several SMILES-based learning approaches \cite{jensen2019graph,gao2022sample,Brown_2019}.

In contrast, REINVENT-Transformer \cite{xu2024reinventtransformer} designs molecules by synthesizing SMILES strings in an autoregressive manner. This method uses a transformer network that is pretrained in a self-supervised fashion on known molecules and then fine-tuned with reinforcement learning via a variant of the REINFORCE algorithm \cite{williams1992simple}. We include REINVENT-Transformer in our comparisons because, like GraphXForm, it relies on pretrained transformers and generates molecules autoregressively. However, while REINVENT-Transformer constructs molecules token by token from a predefined vocabulary, GraphXForm operates directly on the molecular graph by adding atoms and bonds.

\paragraph*{GuacaMol results} 
Table~\ref{tab:Result_Guacamol_OBJ} presents the scores of the best molecules found for four representative \cite{gao2024generative} tasks: Ranolazine MPO, Perindopril MPO, Sitagliptin MPO, and Scaffold Hop. A complete list of results across all 20 tasks is provided in Supplementary Table 1.
For Graph~GA, we report the results as originally published by \citet{Brown_2019}. For REINVENT-Transformer, we conducted experiments using the source code and pretrained model provided by \citet{gao2022sample}. 

GraphXForm outperforms both Graph~GA and REINVENT-Transformer in different drug design tasks.
As shown in Table~\ref{tab:Result_Guacamol_OBJ}, GraphXForm finds molecules with significantly higher scores for the three MPO cases compared to Graph~GA and REINVENT-Transformer.
For the scaffold hop task, GraphXForm achieves the best possible score of 1 similar to Graph~GA.
Considering the perfomance across all 20 tasks of the GuacaMol benchmark (see Supplementary Table 1), GraphXForm attains a total summed score of 18.227, compared to 17.983 for Graph~GA. Furthermore, Supplementary Table 1 demonstrates that our method overall outperforms other classic baselines from the original GuacaMol paper, as well as a recent optimization method that utilizes multiple GPT agents for drug design \cite{hu2024novo}. These results underscore the robust molecular optimization capabilities of GraphXForm.

\begin{table*}[t]
    \centering
    \caption{Performance of different molecular design methods across four tasks from the goal-directed GuacaMol \cite{Brown_2019} benchmark. We report the objective function evaluation of the best molecule found.}
    \label{tab:Result_Guacamol_OBJ}
    \renewcommand{\arraystretch}{1.2} 
    \setlength{\tabcolsep}{6pt} 
    
    \begin{tabular}{lcccc}
        \toprule
        \textbf{Method} & \textbf{Ranolazine MPO} & \textbf{Perindopril MPO} & \textbf{Sitagliptin MPO} & \textbf{Scaffold Hop} \\
        \midrule
        Graph~GA~ \cite{jensen2019graph} & 0.920 & 0.792 & 0.891 & \textbf{1.000} \\
        REINVENT-Transformer~ \cite{xu2024reinventtransformer} & 0.934 & 0.679 & 0.735 & 0.582 \\
        \textbf{GraphXForm (ours)} & \textbf{0.944} & \textbf{0.835} & \textbf{0.965} & \textbf{1.000} \\
        \bottomrule
    \end{tabular}
\end{table*}

\subsection{Case study 2: Solvent design}\label{subsec:design_task_2}
\paragraph*{Objectives}
To further evaluate GraphXForm beyond already-established drug design benchmarks, we propose a solvent design task for two-phase aqueous-organic systems used in liquid-liquid extraction. Our focus is on two examples motivated by biotechnology, where products are typically produced in an aqueous solution using microorganisms or enzymes. In such processes, products are to be extracted using the organic solvents we aim to design. We assume a spatially uniform temperature of 298 K in both examples. 

The first solvent design task focuses on the separation of isobutanol (IBA) from water, a common liquid-liquid extraction process. The chosen solvent should be largely immiscible with water (i.e., low solubility exhibited for both the solvent in water and water in the solvent) and possess high affinity for IBA. As is common practice in chemical engineering, we use the partition coefficient $P_{\text{IBA}}^{\infty}$ at small mole fractions of IBA in both phases $x_\text{IBA}$:
\begin{equation}
    P_{\text{IBA}}^{\infty} = \lim_{x_\text{IBA}^\text{W} \to 0} \frac{x_\text{IBA}^\text{S}}{x_\text{IBA}^\text{W}}
\end{equation}
where $x_\text{IBA}^\text{W}$ and $x_\text{IBA}^\text{S}$ are the mole fractions of IBA in water (W) and the solvent (S), respectively. This coefficient serves as a simple yet effective measure of the relative affinity of the solvent compared to water. Assuming low mutual solubility between the solvent and water, $P_{\text{IBA}}^{\infty}$ can be well approximated by the ratio of IBA's activity coefficients at infinite dilution, $\gamma_\text{IBA,W}^\infty$ and $\gamma_\text{IBA,S}^\infty$, in water and solvent, respectively:
\begin{equation}
    P_{\text{IBA}}^{\infty} = \frac{\gamma_\text{IBA,W}^\infty}{\gamma_\text{IBA,S}^\infty}.
\end{equation}

To ensure the formation of two phases, i.e., a miscibility gap between the solvent and water, we use the following constraint:
\begin{equation}  \label{eq:misc_constraint}
     \gamma_\text{S,W}^\infty \cdot \gamma_\text{W,S}^\infty > \exp(4). 
\end{equation}
This constraint guarantees a phase split between the water and solvent, assuming that the activity coefficient profiles follow the two-parameter Margules $g^E$ model \cite{wisniak1983liquid}. Although the activity coefficients of all conceivable solvent/water mixtures will not necessarily follow this model, the constraint still serves as a useful indicator for miscibility gaps.

The partition coefficient and the miscibility gap constraint are then combined to form the following scalar objective function:

\begin{equation}\label{eq:obj2}
     \max \frac{1}{\gamma_\text{IBA,S}^\infty} + \left(\tanh\left(\gamma_\text{S,W}^\infty \cdot \gamma_\text{W,S}^\infty - \exp(4)\right) - 1 \right) \cdot 10.
\end{equation}
Hereby, the solvent-independent constant $\gamma_\text{IBA,W}^\infty$ is omitted. 

Our second solvent design task centers on an extraction process presented by Peters et al.~ \cite{Peters.2008}, who carried out a solvent screening using COSMO-RS as a property predictor. Here, an enzymatic reaction in aqueous medium converts 3,5-dimethoxy-benzaldehyde (DMBA) molecules to (R)-3,3’,5,5’-tetra-methoxy-benzoin (TMB). The task is to find an organic solvent that forms a two-phase system with water. Similar to the IBA task, an optimal solvent should have a high affinity for the product TMB, enabling it to pull TMB out of the aqueous phase. At the same time, however, the solvent should have a low affinity for the educt DMBA. Designing a suitable solvent for this task is extremely challenging because DMBA and TMB possess similar chemical structures and polarities.

Again assuming small concentrations of DMBA and TMB as well as low mutual solubility between the solvent and water, we define the following partition coefficients similarly to our IBA task:
\begin{equation}
    \begin{aligned}
        P_{\text{DMBA}}^{\infty} = \frac{\gamma_\text{DMBA,W}^\infty}{\gamma_\text{DMBA,S}^\infty} \\
        P_{\text{TMB}}^{\infty} = \frac{\gamma_\text{TMB,W}^\infty}{\gamma_\text{TMB,S}^\infty}. \\
    \end{aligned}
\end{equation}

Following Peters et al.~ \cite{Peters.2008}, we maximize the ratio $P_{\text{TMB}}^{\infty} / P_{\text{DMBA}}^{\infty}$, while additionally enforcing the miscibility gap constraint from Equation \ref{eq:misc_constraint} leading to the following scalar objective:
\begin{equation}\label{eq:obj1}
     \max \frac{\gamma_\text{TMB,S}^\infty}{\gamma_\text{DBMA,S}^\infty} + \left(\tanh\left(\gamma_\text{S,W}^\infty \cdot \gamma_\text{W,S}^\infty - \exp(4)\right) - 1 \right) \cdot 10.
\end{equation}
Hereby, we again omitted the constants $\gamma_\text{TMB,W}^\infty$ and $\gamma_\text{DBMA,W}^\infty$.

We note that further thermodynamic properties, e.g., boiling and melting points, and sustainability indicators, such as biodegradability and toxicity, are highly relevant for the practical effectiveness of solvents.
Such properties could be considered as part of the objective function or as additional constraints in future work.
\paragraph*{Property prediction}
To obtain activity coefficients at infinite dilution, we use a state-of-the-art graph neural network (GNN)~ \cite{Gilmer.2017, Reiser.2022, Rittig_GNNBook.2022}.
Specifically, we employ the Gibbs-Helmholtz (GH-) GNN~ \cite{SanchezMedina2023} that was developed by Sanchez Medina \textit{et al.} for predicting activity coefficients at infinite dilution of binary mixtures at varying temperature.  
The GH-GNN takes the molecular graphs of the two molecules within a binary mixture and the temperature as inputs.
First, structural information from the individual molecular graphs and molecular (self-)interactions based on a mixture graph are encoded into a vector representation, the mixture fingerprint.
Based on the mixture fingerprint, a multilayer perceptron (MLP) then predicts the parameters of the Gibbs-Helmholtz relationship so that infinite dilution activity coefficients can be predicted with temperature.
The GH-GNN is trained in a structure-to-property manner, i.e., it directly learns activity coefficients at infinite dilution as a function of the molecular graphs. A large data set of experimental activity coefficients at infinite dilution from the DECHEMA Chemistry Data Series~ \cite{gmehling2008dechema} was used for training. For further details on the GH-GNN architecture, we refer to~ \cite{SanchezMedina2023}.

We note that many GNN models and other ML models such as transformers have been developed for activity coefficient prediction~ \cite{Rittig2023a, Winter.2022, Damay.2021}, also considering the composition-dependency and thermodynamic consistency~ \cite{Rittig2023b, Winter.2023, Rittig.2024, specht2024hanna}. 
We here chose the GH-GNN as it is specialized for activity coefficients at infinite dilution and was trained on a much larger experimental database than the other models, thus covering a larger chemical space~ \cite{SanchezMedina2023}, which is desirable for molecular design.
This model has shown high prediction accuracy, outperforming well-established methods for predicting activity coefficients at infinite dilution such as UNIFAC~ \cite{Fredenslund.1975} or COSMO-RS~ \cite{Klamt.2010}, cf. \citet{SanchezMedina2023}.
Future work could investigate further additional activity coefficient models or directly predicting partition coefficients with ML~ \cite{Zamora2023, nevolianis2024multi}.

\paragraph*{Alphabet and size constraints}
We focus our solvent design task on organic H-C-N-O chemistry, restricting the available alphabet to $\Sigma = (\mathrm C, \mathrm N, \mathrm O)$ without ionization and chirality. This choice emphasizes the primary building blocks and reduces the likelihood of designing unstable and unsustainable solvents by omitting elements more likely to contribute to these issues. Furthermore, to prevent potential exploitation of the surrogate by generating excessively large molecules that exceed the typical size of solvents, we constrain the design to molecules with no more than 25 atoms.

\paragraph*{Benchmark methods} 
In addition to Graph~GA and REINVENT-Transformer, we benchmark GraphXForm for solvent design against STONED \cite{nigam2021beyond} and the Junction Tree Variational Autoencoder (JT-VAE) \cite{Jin2018}. For these methods, we use the source code provided by \citet{gao2022sample}. STONED is a simple yet efficient algorithm that employs a GA operating at the string level, manipulating tokens within the SELFIES molecular representation \cite{krenn2020self}.

JT-VAE adopts a VAE, a widely used generative model in ML-guided molecular design \cite{sanchez2018inverse, bilodeau2022generative, anstine2023generative}.
VAEs use an encoder-decoder structure, where the encoder maps molecules into a continuous latent space, and the decoder reconstructs them from this representation. 
Particularly when the objective function is derived from a trained network, the molecular latent space can facilitate exploration of the molecular space.
That is, different optimization strategies can be employed to discover points in the latent space that correspond to promising novel molecules. 
These strategies include random sampling, Bayesian optimization, and GAs \cite{sanchez2018inverse, rittig2023graph, anstine2023generative}. In our work, we use JT-VAE in combination with GAs. JT-VAE operates on molecular graphs and their non-cyclic abstractions (junction trees), and it has demonstrated a high rate of decoding latent vectors into chemically valid molecules. Since we consider only molecules that conform to the alphabet $\Sigma$, we train JT-VAE on a subset of the QM9 dataset \cite{ruddigkeit2012enumeration,ramakrishnan2014quantum} consisting of approximately 128,000 molecules with at most nine heavy atoms.

\subsubsection{Unconstrained results}\label{subsec:unconstrained_results}

The goal for all methods is to find suitable solvents with respect to the two objectives from Equations~\ref{eq:obj2} and \ref{eq:obj1} for the two example problems, using the general setup outlined in Section~\ref{subsec:design_task_2}.
In addition to the molecules shown in this section, a list of the top 20 molecules from the best runs can be found in Supplementary Figures~1-14.

\begin{table*}[t]
    \centering
    \caption{Performance of different molecular design methods for the two example solvent design tasks. Each method is run across three different random seeds, with a maximum time budget of 8 hours. We report the objective function evaluation of the best molecule found over all runs (`max best'), best averaged over all the three runs $\pm$ standard deviation (`mean best'), the average of the top 20 molecules of the best run (`max top 20'), and the mean of the top 20 over all three runs $\pm$ standard deviation (`mean top 20').}
    \label{tab:Result_Design_OBJ}%
    \resizebox{\linewidth}{!}{%
    \begin{tabular}{l|cccc|cccc}
        \toprule
        \multicolumn{1}{c|}{\multirow{2}[3]{*}{\shortstack[c]{Method}}} & \multicolumn{4}{c|}{IBA (cf. (\ref{eq:obj2}))} & \multicolumn{4}{c}{TMB/DMBA (cf. (\ref{eq:obj1}))} \\
         & max best & mean best & max top 20 & mean top 20 &  max best & mean best & max top 20 & mean top 20 \\
        \cmidrule(lr){1-1}\cmidrule(lr){2-5}\cmidrule(lr){6-9} 

            JT-VAE \cite{Jin2018}                & 6.85 & 6.41 $\pm$ 0.66 & 6.04 & 5.68 $\pm$ 0.57& 2.16 & 1.56 $\pm$ 0.54 & 1.44 & 1.20 $\pm$ 0.23   \\
            STONED \cite{nigam2021beyond}                & 8.31 & 7.42 $\pm$ 0.94 & 6.72 & 6.28 $\pm$ 0.41 & 2.39 & 1.68 $\pm$ 0.65 & 1.91 & 1.45 $\pm$ 0.49 \\
            Graph~GA \cite{jensen2019graph}               & \textbf{8.87} & 7.13 $\pm$ 3.01 & {\bf 8.67} & 6.80 $\pm$ 3.22 & 8.40 & 8.14 $\pm$ 0.27 & 8.07 & 7.95 $\pm$ 0.32 \\
            REINVENT-Transformer \cite{xu2024reinventtransformer}    & \textbf{8.87} & 8.32 $\pm$ 0.52 & 8.66 & 8.09 $\pm$ 0.45 & 7.41 & 6.52 $\pm$ 1.17 & 7.22 & 6.42 $\pm$ 0.96  \\
            \textbf{GraphXForm (ours)}       & \textbf{8.87} & \textbf{8.87 $\pm$ 0.00} & 8.60 & \textbf{8.58 $\pm$ 0.04} & \textbf{8.65} & \textbf{8.65 $\pm$ 0.00} & \textbf{8.41} & \textbf{8.39 $\pm$ 0.01}  \\
        \bottomrule
    \end{tabular}%
    }
\end{table*}

\paragraph*{Screening list} 

\begin{figure*}
    \center
  \includegraphics[width=0.9\linewidth]{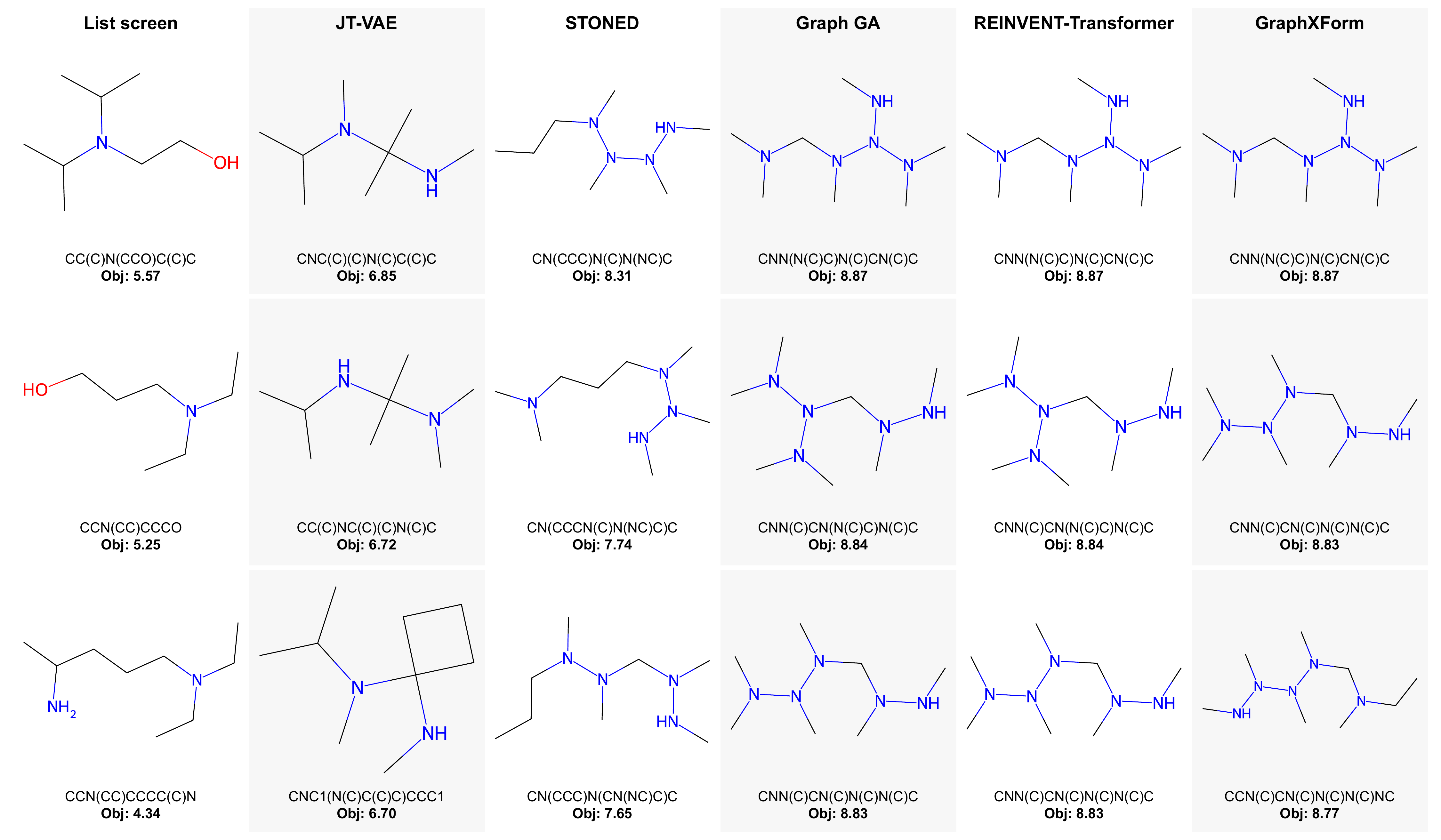}
  \caption{{\bf IBA task, unconstrained}: Top three molecules (with their corresponding SMILES string and objective value) identified by each method across all runs.}
  \label{fig:results_obj1}
\end{figure*}

\begin{figure*}
    \center
  \includegraphics[width=0.9\linewidth]{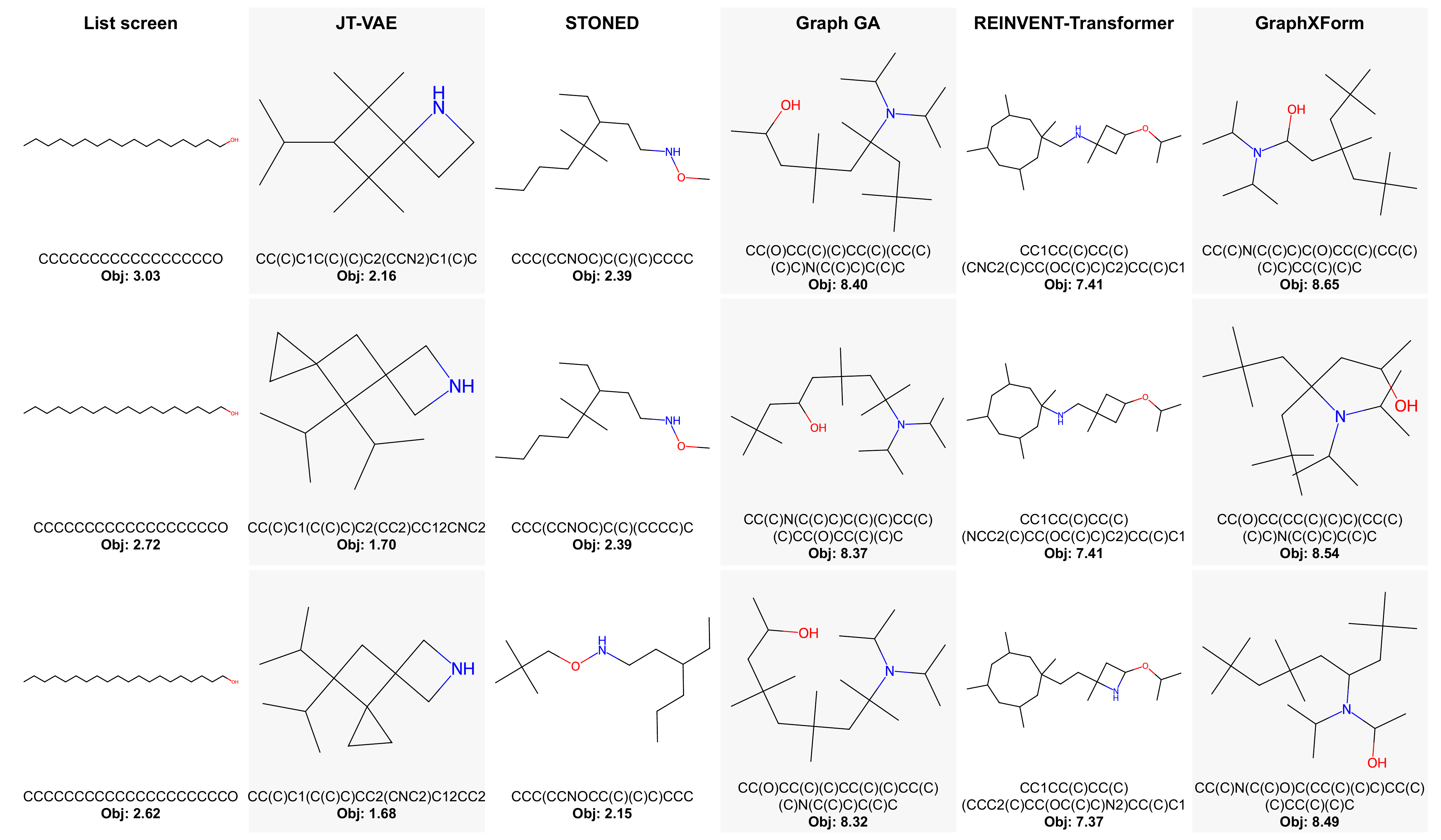}
  \caption{{\bf TMB/DMBA task, unconstrained}: Top three molecules (with their corresponding SMILES string and objective value) identified by each method across all runs.}
  \label{fig:results_obj2}
\end{figure*}

To contextualize the results of all methods, we selected all 6,098 compounds from the COSMObase 2020 database that conform to the alphabet $\Sigma$. We calculated the objective functions for those compounds that also meet the miscibility gap requirements, see Eq. (\ref{eq:misc_constraint}). The top three compounds based on their objective values are shown in the first column in Figures~\ref{fig:results_obj1} and \ref{fig:results_obj2}. For the IBA task, the best molecule in this list achieves an objective value of 5.57, while for the TMB/DMBA task, the highest objective value is 3.03.

\paragraph*{Model results} 

Table~\ref{tab:Result_Design_OBJ} compares the performance of the different molecular design methods for the two solvent design tasks.
We report the objective value of the best molecule found by each method, as well as the average objective values of the top 20 molecules. We also present results averaged over multiple runs with different random seeds, providing insight into the robustness of each method. Additionally, Figures~\ref{fig:results_obj1} and \ref{fig:results_obj2} display the structural formulas of the top three molecules identified by each method. We note that, in these results, we do not yet impose any structural constraints and focus solely on optimization.

In the IBA task, GraphXForm, REINVENT-Transformer, and Graph~GA all identified the same best molecule, which has an objective value of 8.87. However, GraphXForm consistently found this molecule in every run, highlighting its stability. For example, in contrast, while Graph~GA has a slightly higher average value for the best 20 molecules found, its mean best objective over all runs is only 7.13.

For the more challenging TMB/DMBA task, GraphXForm outperforms all other methods across every metric. Notably, Graph~GA outperforms REINVENT-Transformer overall, likely due to the latter's sensitivity to initialization during RL fine-tuning. This factor is also reflected in its relatively high standard deviation. We also observe that the designs produced by Graph~GA and GraphXForm exhibit substantial structural similarity, although GraphXForm makes additional refinements that lead to improved scores.

Interestingly, the JT-VAE and STONED methods identified molecules with significantly lower objective values for the TMB/DMBA task when compared to other methods, with maximum values of approximately 2.16 and 2.39, respectively. However, their results for the IBA task (6.85 and 7.53) were closer to those of the other methods. We attribute this discrepancy to the nature of the tasks: The TMB/DMBA task is inherently more challenging and requires larger molecules with more complex branching, while the best-performing molecules in the IBA task were smaller. As JT-VAE was trained only on molecules with up to nine heavy atoms, its ability to explore larger molecules seems limited. Similarly, STONED struggled to effectively explore larger molecular structures.

In summary, GraphXForm demonstrates highly promising results in both solvent design tasks, outperforming its comparison partners in terms of identifying the best candidates and ensuring robustness in the design process. In the following sections, we further explore the flexibility of GraphXForm by imposing structural constraints on the designed molecules and starting the design process from initial molecular structures.

\subsubsection{Including structural constraints in molecular design}\label{subsec:constrained_results}

In addition to optimizing an objective, such as a physical property, it is often desirable to enforce specific structural characteristics in the generated molecular candidates. These structural constraints can include limitations on ring size or specific bonding patterns between atom types, which can improve chemical feasibility, such as enhancing synthesizability and stability or reducing toxic moieties. For example, in the IBA task, although the top molecules generated by GraphXForm achieve high objective values, they exhibit features that pose challenges for solvent design. Notably, the presence of single nitrogen–nitrogen bonds and single nitrogen–carbon–nitrogen bonds may lead to instability in water \cite{hydrzine}, resulting in reactive compounds unsuitable for liquid–liquid extraction. Similarly, our top result for the TMB/DMBA task includes a carbon atom that is single-bonded to a nitrogen, a hydroxyl group, and a single hydrogen atom, which also raises concerns about instability.

A straightforward approach to enforcing such constraints is to assign an objective value of $-\infty$ to any designed molecules that violate them, effectively discarding those candidates. In contrast, because GraphXForm operates directly on the molecular graph – transitioning from one molecule to the next without waiting for the completion of a string – we can flexibly restrict the search space by simply masking actions that would lead to constraint violations.

To examine how structural constraints affect the performance of GraphXForm in finding molecules with high objective values, we impose the following constraints: rings are limited to five or six atoms, and the following bonding patterns are disallowed: single bonds between two nitrogen atoms; single bonds between two oxygen atoms; a carbon atom single-bonded to two nitrogen atoms, unless the carbon is also double-bonded to an oxygen atom, forming a urea functional group; and a carbon atom single-bonded to a nitrogen atom, an oxygen atom, a hydrogen atom, and one other non-hydrogen functional group. We acknowledge that these constraints do not cover all aspects of synthetic accessibility, chemical stability, safety, or environmental impact. Rather, they serve as illustrative examples to demonstrate the capabilities of GraphXForm in incrementally limiting the designs to promising candidate structures that can be further evaluated by experts.

\begin{figure}
    \center
  \includegraphics[width=\linewidth]{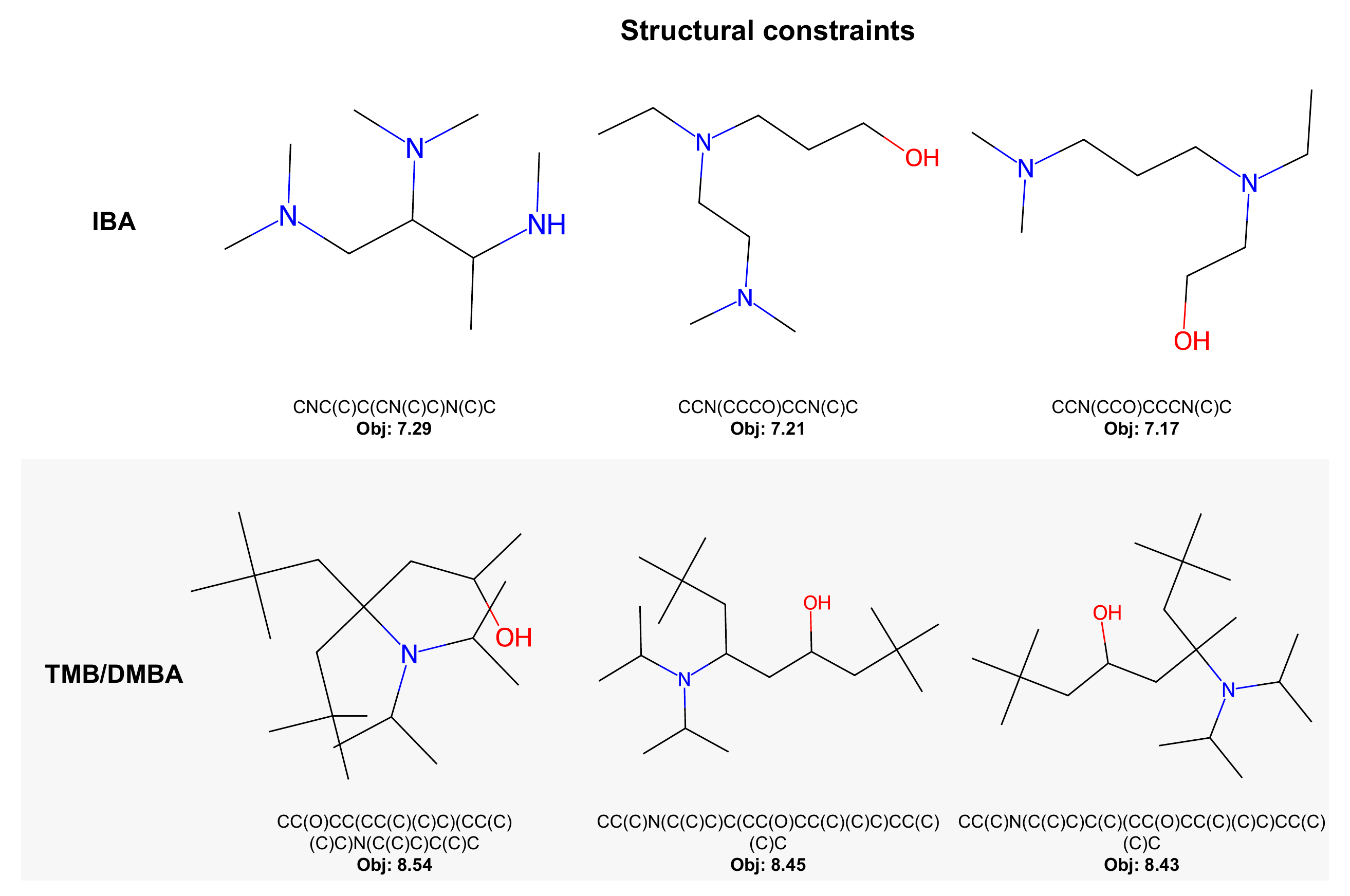}
  \caption{Top three molecules designed by GraphXForm under structural constraints on specific bonding patterns and ring sizes.}
  \label{fig:results_strcon}
\end{figure}

For the IBA task, the top three molecules designed by GraphXForm under these constraints are shown in the top row of Figure~\ref{fig:results_strcon}. With constraints in place, we achieve a `mean best' value of $7.28 \pm 0.00$ and a `mean top 20' value $6.85 \pm 0.02$, which are only slightly lower than the values obtained without structural constraints. We also conducted constrained design runs with REINVENT-Transformer as a string-based counterpart; however, it reached only a ‘mean best’ value of $6.36 \pm 0.39$, with its overall best molecule scoring $6.92$. 

The bottom row of Figure~\ref{fig:results_strcon} displays the top three molecules identified by GraphXForm under the structural constraints for the TMB/DMBA task. Under these constraints, GraphXForm achieves a `mean best' value of $8.54 \pm 0.00$ and a `mean top 20' value of $8.31 \pm 0.06$. Notably, these results remain superior to those of the other methods without structural constraints, as reported in Table~\ref{tab:Result_Design_OBJ}.

These findings further demonstrate GraphXForm’s flexibility and its ability to consistently design promising molecules.

\subsubsection{Starting molecule design from initial structures}

\begin{figure}
    \center
  \includegraphics[width=\linewidth]{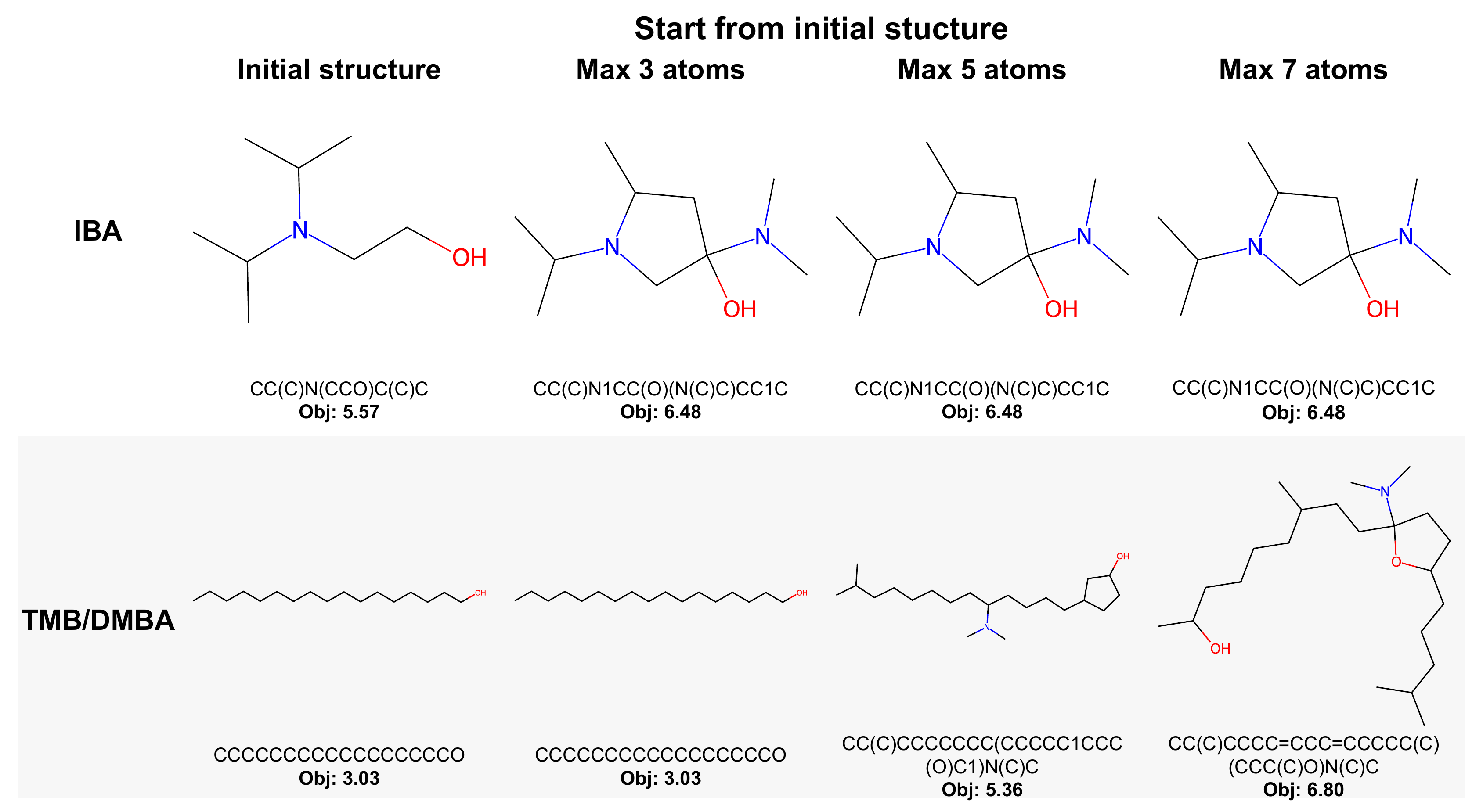}
  \caption{Best molecule designed by GraphXForm when augmenting a specific initial structure by adding up to 3, 5, or 7 atoms. Bonds can be added without restriction. For the TMB/DMBA task, the designed molecule is also required to remain an alcohol.}
  \label{fig:results_startstr}
\end{figure}

One advantage of our molecular graph approach, which is not achievable out of the box with vanilla SMILES-based next-token prediction due to its linear construction, is the direct ability to initiate the design process from a predefined structure. This feature is especially useful when a known candidate, which already possesses desirable properties, could benefit from targeted modifications to further improve the objective function. To demonstrate this, we consider the IBA task and start with the best molecule from COSMObase 2020, which has an objective value of 5.57 (see Figure~\ref{fig:results_obj2}). We then conduct three separate experimental runs, allowing GraphXForm to add up to 3, 5, and 7 atoms to the initial structure while adding bonds as needed, all under the structural constraints outlined in Section~\ref{subsec:constrained_results}. The original molecule and the best molecule from each of the three runs are shown in the top row of Figure~\ref{fig:results_startstr}. In every case, the same modified molecule is produced (with three additional atoms), resulting in an improved objective value.

We repeat a similar experiment for the TMB/DMBA task. During list screening, we observed that the top three structures were all long-chain alcohols. However, long-chain alcohols tend to have melting points above room temperature, making them unsuitable as solvents for the intended processes. The melting point can be lowered by, e.g., adding branches. To address this, we again initiated three runs starting from the best molecule identified in the screening. These runs allowed the addition of up to 3, 5, and 7 atoms, thereby simulating branching of the long-chain alcohol. Additionally, we constrained the design process so that the resulting molecule remains an alcohol by preventing any modifications to the hydroxy group (an oxygen atom bonded to one hydrogen atom). The results, shown in the second row of Figure~\ref{fig:results_startstr}, indicate that while no improvement in the objective function is observed when adding only 3 atoms, a notable enhancement occurs when 5 or 7 atoms are added.

This approach provides a powerful tool for cases where it is preferable to build upon an existing molecule rather than starting from scratch. We note that enabling the agent to remove atoms or bonds would not be beneficial when designing from a single atom, as it would introduce unnecessary redundancies into the search space. However, in scenarios where the design process begins with an existing molecule, allowing removal could offer additional flexibility and enable greater deviations from the initial structure. Extending the action space to include atom and bond removal is straightforward within our framework; however, we leave the exploration of this possibility for future work.

\section*{Conclusion}\label{sec:conclusion}
We presented GraphXForm, a method for molecular design that follows the successful paradigm of self-supervised pretraining followed by (RL-based) fine-tuning, but operates directly on molecular graphs. By doing so, we addressed challenges faced by string-based methods, such as chemical validity or accomodating structural constraints. We introduced a technique derived from self-improvement learning to fine-tune a deep graph-transformer model. Based on the established drug design GuacaMol benchmark \cite{Brown_2019} and two solvent design tasks, we showed that GraphXForm can outperform state-of-the-art molecule design techniques. Additionally, our approach can flexibly adhere to specified structural constraints, such as bond types and functional groups, and can adaptively start the design process from existing molecular structures. 

Looking ahead, several promising avenues for future development exist. First, we plan to expand the atom alphabet of GraphXForm and pretrain the network on significantly larger databases. Although these enhancements can be integrated into our current framework, we intentionally limited the atom set in this study to ensure comparability with other methods. Second, we aim to extend the action space by allowing the agent to {\em remove} bonds and atoms. While not necessary for our present investigation – and admittedly introducing further action symmetries – this addition would offer the model more possibilities when modifying starting molecules.

Additionally, future work could involve evaluating GraphXForm on a broader range of molecular design tasks and incorporating more design considerations. For instance, recent trends in ML-based drug development emphasize increased sample efficiency \cite{gao2022sample,guo2024augmented, guo2024saturn, rittig2024deterministic} and the integration of synthesizability models \cite{guo2024directly, gao2024generative}. Although sample efficiency was not the primary focus of our current study, our method inherently maintains a kind of experience replay buffer, suggesting that techniques targeting replay buffers \cite{guo2024augmented,guo2024saturn} could be readily applied to further enhance GraphXForm. Furthermore, tweaking TASAR parameters (e.g., using a shallower beam width $\beta$ or a higher step size $\sigma$) and allowing more training batches in each fine-tuning epoch can help to make the method more sample-efficient.

We also aim to incorporate more constraints to ensure the suitability of the designed molecules for specific applications.
In the case of solvents for liquid-liquid separation processes, this includes recognizing that numerous factors – such as boiling and melting points – are critical to a solvent's effectiveness. 
This, however, will depend on the availability of reliable property predictors. 

Finally, since many structural constraints (e.g., presence of certain atoms, bonds, or formal groups) on the molecular graph can be flexibly formulated and implemented in a general manner within the current framework, we envision integrating GraphXForm with large language models to create a user-friendly design interface. This would allow researchers to formulate constraints in natural language, which would then be translated into the appropriate configuration for the model.

\section*{Author contributions}
J.P.: Conceptualization; Methodology - Concept and implementation of environment, network architecture, learning algorithm, adaptation of benchmark methods; Analysis; Validation; Visualization; Writing - Original Draft. J.G.R.: Conceptualization; Methodology - Concept and implementation of objective functions and surrogate models, adaptation of benchmark methods; Analysis; Validation; Writing - Original Draft. A.B.W.: Conceptualization; Methodology - Concept and implementation of environment and objective functions, adaptation of benchmark methods; Analysis; Validation; Visualization; Writing - Original Draft. M.G.: Resources; Funding acquisition; Writing - Review \& Editing. J.B.: Conceptualization; Resources; Supervision; Project administration; Funding acquisition; Writing - Review \& Editing. A.M.: Conceptualization; Resources; Supervision; Project administration; Funding acquisition; Writing - Review \& Editing. D.G.G.: Conceptualization; Resources; Supervision; Project administration; Funding acquisition; Writing - Review \& Editing.

\section*{Data availability}

All our code and data for pretraining and fine-tuning GraphXForm is available at \url{https://github.com/grimmlab/graphxform}.

\section*{Acknowledgements}

This project was funded by the Deutsche Forschungsgemeinschaft (DFG, German Research Foundation) – 466417970 and 466387255 – within the Priority Programme ``SPP 2331: Machine Learning in Chemical Engineering''.
This work was performed as part of the Helmholtz School for Data Science in Life, Earth and Energy (HDS-LEE).
Simulations were performed with computing resources granted by RWTH Aachen University under project ``thes1232''. 
The authors gratefully acknowledge the Competence Center for Digital Agriculture (KoDA) at the University of Applied Sciences Weihenstephan-Triesdorf for providing additional computational resources. Furthermore, the authors thank Herbert Riepl for suggesting structural constraints.

\bibliography{main}

\begin{thebibliography}{89}
\providecommand{\natexlab}[1]{#1}
\providecommand{\url}[1]{\texttt{#1}}
\expandafter\ifx\csname urlstyle\endcsname\relax
  \providecommand{\doi}[1]{doi: #1}\else
  \providecommand{\doi}{doi: \begingroup \urlstyle{rm}\Url}\fi

\bibitem[hyd(2024)]{hydrzine}
National center for biotechnology information. pubchem compound database, 2024.
\newblock URL \url{https://pubchem.ncbi.nlm.nih.gov/}.

\bibitem[Anstine \& Isayev(2023)Anstine and Isayev]{anstine2023generative}
Anstine, D.~M. and Isayev, O.
\newblock Generative models as an emerging paradigm in the chemical sciences.
\newblock \emph{Journal of the American Chemical Society}, 145\penalty0
  (16):\penalty0 8736--8750, 2023.

\bibitem[Anthony et~al.(2017)Anthony, Tian, and Barber]{anthony2017thinking}
Anthony, T., Tian, Z., and Barber, D.
\newblock Thinking fast and slow with deep learning and tree search.
\newblock \emph{Advances in neural information processing systems}, 30, 2017.

\bibitem[Ar{\'u}s-Pous et~al.(2020)Ar{\'u}s-Pous, Patronov, Bjerrum, Tyrchan,
  Reymond, Chen, and Engkvist]{arus2020smiles}
Ar{\'u}s-Pous, J., Patronov, A., Bjerrum, E.~J., Tyrchan, C., Reymond, J.-L.,
  Chen, H., and Engkvist, O.
\newblock Smiles-based deep generative scaffold decorator for de-novo drug
  design.
\newblock \emph{Journal of cheminformatics}, 12:\penalty0 1--18, 2020.

\bibitem[Bachlechner et~al.(2021)Bachlechner, Majumder, Mao, Cottrell, and
  McAuley]{bachlechner2021rezero}
Bachlechner, T., Majumder, B.~P., Mao, H., Cottrell, G., and McAuley, J.
\newblock Rezero is all you need: Fast convergence at large depth.
\newblock In \emph{Uncertainty in Artificial Intelligence}, pp.\  1352--1361.
  PMLR, 2021.

\bibitem[Bilodeau et~al.(2022)Bilodeau, Jin, Jaakkola, Barzilay, and
  Jensen]{bilodeau2022generative}
Bilodeau, C., Jin, W., Jaakkola, T., Barzilay, R., and Jensen, K.~F.
\newblock Generative models for molecular discovery: Recent advances and
  challenges.
\newblock \emph{Wiley Interdisciplinary Reviews: Computational Molecular
  Science}, 12\penalty0 (5):\penalty0 e1608, 2022.

\bibitem[Bjerrum \& Threlfall(2017)Bjerrum and
  Threlfall]{Bjerrum2017MolecularGW}
Bjerrum, E.~J. and Threlfall, R.
\newblock Molecular generation with recurrent neural networks (rnns).
\newblock \emph{arXiv preprint arXiv:1705.04612}, 2017.

\bibitem[Brown et~al.(2019)Brown, Fiscato, Segler, and Vaucher]{Brown_2019}
Brown, N., Fiscato, M., Segler, M.~H., and Vaucher, A.~C.
\newblock Guacamol: Benchmarking models for de novo molecular design.
\newblock \emph{Journal of Chemical Information and Modeling}, 59\penalty0
  (3):\penalty0 1096--1108, 2019.
\newblock \doi{10.1021/acs.jcim.8b00839}.
\newblock URL \url{https://doi.org/10.1021/acs.jcim.8b00839}.
\newblock PMID: 30887799.

\bibitem[Cheng et~al.(2023)Cheng, Cai, Miret, Malkomes, Phielipp, and
  Aspuru-Guzik]{cheng2023group}
Cheng, A.~H., Cai, A., Miret, S., Malkomes, G., Phielipp, M., and Aspuru-Guzik,
  A.
\newblock Group selfies: a robust fragment-based molecular string
  representation.
\newblock \emph{Digital Discovery}, 2\penalty0 (3):\penalty0 748--758, 2023.

\bibitem[Corsini et~al.(2024)Corsini, Porrello, Calderara, and
  Dell'Amico]{corsini2024self}
Corsini, A., Porrello, A., Calderara, S., and Dell'Amico, M.
\newblock Self-labeling the job shop scheduling problem.
\newblock In \emph{The Thirty-eighth Annual Conference on Neural Information
  Processing Systems}, 2024.
\newblock URL \url{https://openreview.net/forum?id=buqvMT3B4k}.

\bibitem[Dai et~al.(2018)Dai, Tian, Dai, Skiena, and
  Song]{dai2018syntaxdirected}
Dai, H., Tian, Y., Dai, B., Skiena, S., and Song, L.
\newblock Syntax-directed variational autoencoder for structured data.
\newblock In \emph{International Conference on Learning Representations}, 2018.
\newblock URL \url{https://openreview.net/forum?id=SyqShMZRb}.

\bibitem[Damay et~al.(2021)Damay, Jirasek, Kloft, Bortz, and Hasse]{Damay.2021}
Damay, J., Jirasek, F., Kloft, M., Bortz, M., and Hasse, H.
\newblock Predicting activity coefficients at infinite dilution for varying
  temperatures by matrix completion.
\newblock \emph{Industrial {\&} Engineering Chemistry Research}, 60\penalty0
  (40):\penalty0 14564--14578, 2021.
\newblock ISSN 0888-5885.
\newblock \doi{10.1021/acs.iecr.1c02039}.

\bibitem[Davies et~al.(2015)Davies, Nowotka, Papadatos, Dedman, Gaulton,
  Atkinson, Bellis, and Overington]{davies2015chembl}
Davies, M., Nowotka, M., Papadatos, G., Dedman, N., Gaulton, A., Atkinson, F.,
  Bellis, L., and Overington, J.~P.
\newblock Chembl web services: streamlining access to drug discovery data and
  utilities.
\newblock \emph{Nucleic acids research}, 43\penalty0 (W1):\penalty0 W612--W620,
  2015.

\bibitem[De~Cao \& Kipf(2018)De~Cao and Kipf]{de2018molgan}
De~Cao, N. and Kipf, T.
\newblock Molgan: An implicit generative model for small molecular graphs.
\newblock \emph{arXiv preprint arXiv:1805.11973}, 2018.

\bibitem[Devlin et~al.(2019)Devlin, Chang, Lee, and Toutanova]{devlin2018bert}
Devlin, J., Chang, M.-W., Lee, K., and Toutanova, K.
\newblock Bert: Pre-training of deep bidirectional transformers for language
  understanding.
\newblock In \emph{Proceedings of the 2019 conference of the North American
  chapter of the association for computational linguistics: human language
  technologies, volume 1 (long and short papers)}, pp.\  4171--4186, 2019.

\bibitem[Fredenslund et~al.(1975)Fredenslund, Jones, and
  Prausnitz]{Fredenslund.1975}
Fredenslund, A., Jones, R.~L., and Prausnitz, J.~M.
\newblock Group-contribution estimation of activity coefficients in nonideal
  liquid mixtures.
\newblock \emph{AIChE Journal}, 21\penalty0 (6):\penalty0 1086--1099, 1975.
\newblock ISSN 00011541.
\newblock \doi{10.1002/aic.690210607}.

\bibitem[Fu et~al.(2022)Fu, Gao, Coley, and Sun]{fu2022reinforced}
Fu, T., Gao, W., Coley, C., and Sun, J.
\newblock Reinforced genetic algorithm for structure-based drug design.
\newblock \emph{Advances in Neural Information Processing Systems},
  35:\penalty0 12325--12338, 2022.

\bibitem[Gao et~al.(2022)Gao, Fu, Sun, and Coley]{gao2022sample}
Gao, W., Fu, T., Sun, J., and Coley, C.
\newblock Sample efficiency matters: a benchmark for practical molecular
  optimization.
\newblock \emph{Advances in neural information processing systems},
  35:\penalty0 21342--21357, 2022.

\bibitem[Gao et~al.(2024)Gao, Luo, and Coley]{gao2024generative}
Gao, W., Luo, S., and Coley, C.~W.
\newblock Generative artificial intelligence for navigating synthesizable
  chemical space.
\newblock \emph{arXiv preprint arXiv:2410.03494}, 2024.

\bibitem[Gilmer et~al.(2017)Gilmer, Schoenholz, Riley, Vinyals, and
  Dahl]{Gilmer.2017}
Gilmer, J., Schoenholz, S.~S., Riley, P.~F., Vinyals, O., and Dahl, G.~E.
\newblock {Neural message passing for quantum chemistry}.
\newblock \emph{34th International Conference on Machine Learning, ICML 2017},
  3:\penalty0 2053--2070, 2017.

\bibitem[Gmehling et~al.(2008)Gmehling, Tiegs, Medina, Soares, Bastos, Alessi,
  Kikic, Schiller, and Menke]{gmehling2008dechema}
Gmehling, J., Tiegs, D., Medina, A., Soares, M., Bastos, J., Alessi, P., Kikic,
  I., Schiller, M., and Menke, J.
\newblock Dechema chemisry data series, volume ix activity coefficients at
  infinite dilution.
\newblock \emph{DECHEMA Chemistry Data Series}, 9, 2008.

\bibitem[G{\'o}mez-Bombarelli et~al.(2018)G{\'o}mez-Bombarelli, Wei, Duvenaud,
  Hern{\'a}ndez-Lobato, S{\'a}nchez-Lengeling, Sheberla, Aguilera-Iparraguirre,
  Hirzel, Adams, and Aspuru-Guzik]{Gomez-Bombarelli2016SmilesVAE}
G{\'o}mez-Bombarelli, R., Wei, J.~N., Duvenaud, D., Hern{\'a}ndez-Lobato,
  J.~M., S{\'a}nchez-Lengeling, B., Sheberla, D., Aguilera-Iparraguirre, J.,
  Hirzel, T.~D., Adams, R.~P., and Aspuru-Guzik, A.
\newblock Automatic chemical design using a data-driven continuous
  representation of molecules.
\newblock \emph{ACS central science}, 4\penalty0 (2):\penalty0 268--276, 2018.

\bibitem[Guo \& Schwaller(2024{\natexlab{a}})Guo and
  Schwaller]{guo2024augmented}
Guo, J. and Schwaller, P.
\newblock Augmented memory: Sample-efficient generative molecular design with
  reinforcement learning.
\newblock \emph{Jacs Au}, 4\penalty0 (6):\penalty0 2160--2172,
  2024{\natexlab{a}}.

\bibitem[Guo \& Schwaller(2024{\natexlab{b}})Guo and
  Schwaller]{guo2024directly}
Guo, J. and Schwaller, P.
\newblock Directly optimizing for synthesizability in generative molecular
  design using retrosynthesis models.
\newblock In \emph{AI for Accelerated Materials Design - NeurIPS 2024},
  2024{\natexlab{b}}.
\newblock URL \url{https://openreview.net/forum?id=J63EUbjhSw}.

\bibitem[Guo \& Schwaller(2024{\natexlab{c}})Guo and Schwaller]{guo2024saturn}
Guo, J. and Schwaller, P.
\newblock Saturn: Sample-efficient generative molecular design using memory
  manipulation.
\newblock In \emph{NeurIPS 2024 Workshop on AI for New Drug Modalities},
  2024{\natexlab{c}}.
\newblock URL \url{https://openreview.net/forum?id=bm3aARS2Nw}.

\bibitem[Gupta et~al.(2018)Gupta, M{\"u}ller, Huisman, Fuchs, Schneider, and
  Schneider]{gupta2018generative}
Gupta, A., M{\"u}ller, A.~T., Huisman, B.~J., Fuchs, J.~A., Schneider, P., and
  Schneider, G.
\newblock Generative recurrent networks for de novo drug design.
\newblock \emph{Molecular informatics}, 37\penalty0 (1-2):\penalty0 1700111,
  2018.

\bibitem[Henderson et~al.(2018)Henderson, Islam, Bachman, Pineau, Precup, and
  Meger]{henderson2018deep}
Henderson, P., Islam, R., Bachman, P., Pineau, J., Precup, D., and Meger, D.
\newblock Deep reinforcement learning that matters.
\newblock In \emph{Proceedings of the AAAI conference on artificial
  intelligence}, volume~32, 2018.

\bibitem[Hu et~al.(2024)Hu, Liu, Zhao, and Zhang]{hu2024novo}
Hu, X., Liu, G., Zhao, Y., and Zhang, H.
\newblock De novo drug design using reinforcement learning with multiple gpt
  agents.
\newblock \emph{Advances in Neural Information Processing Systems}, 36, 2024.

\bibitem[Huang et~al.(2023)Huang, Gu, Hou, Wu, Wang, Yu, and
  Han]{huang2023large}
Huang, J., Gu, S.~S., Hou, L., Wu, Y., Wang, X., Yu, H., and Han, J.
\newblock Large language models can self-improve.
\newblock In \emph{The 2023 Conference on Empirical Methods in Natural Language
  Processing}, 2023.
\newblock URL \url{https://openreview.net/forum?id=uuUQraD4XX}.

\bibitem[Jensen(2019)]{jensen2019graph}
Jensen, J.~H.
\newblock A graph-based genetic algorithm and generative model/monte carlo tree
  search for the exploration of chemical space.
\newblock \emph{Chemical science}, 10\penalty0 (12):\penalty0 3567--3572, 2019.

\bibitem[Jiang et~al.(2024)Jiang, Dieng, and Webb]{jiang2024property}
Jiang, S., Dieng, A.~B., and Webb, M.~A.
\newblock Property-guided generation of complex polymer topologies using
  variational autoencoders.
\newblock \emph{npj Computational Materials}, 10\penalty0 (1):\penalty0 139,
  2024.

\bibitem[Jin et~al.(2018)Jin, Barzilay, and Jaakkola]{Jin2018}
Jin, W., Barzilay, R., and Jaakkola, T.
\newblock {Junction tree variational autoencoder for molecular graph
  generation}.
\newblock \emph{35th International Conference on Machine Learning, ICML 2018},
  5:\penalty0 3632--3648, 2018.

\bibitem[Klamt et~al.(2010)Klamt, Eckert, and Arlt]{Klamt.2010}
Klamt, A., Eckert, F., and Arlt, W.
\newblock {COSMO-RS: An Alternative to Simulation for Calculating Thermodynamic
  Properties of Liquid Mixtures}.
\newblock \emph{Annual Review of Chemical and Biomolecular Engineering},
  1\penalty0 (1):\penalty0 101--122, jun 2010.
\newblock ISSN 1947-5438.
\newblock \doi{10.1146/annurev-chembioeng-073009-100903}.

\bibitem[K{\"o}nig-Mattern et~al.(2024)K{\"o}nig-Mattern, Medina, Komarova,
  Linke, Rihko-Struckmann, Luterbacher, and Sundmacher]{konig2024machine}
K{\"o}nig-Mattern, L., Medina, E. I.~S., Komarova, A.~O., Linke, S.,
  Rihko-Struckmann, L., Luterbacher, J.~S., and Sundmacher, K.
\newblock Machine learning-supported solvent design for lignin-first
  biorefineries and lignin upgrading.
\newblock \emph{Chemical Engineering Journal}, 495:\penalty0 153524, 2024.

\bibitem[Kool et~al.(2019)Kool, Van~Hoof, and Welling]{kool2019stochastic}
Kool, W., Van~Hoof, H., and Welling, M.
\newblock Stochastic beams and where to find them: The gumbel-top-k trick for
  sampling sequences without replacement.
\newblock In \emph{International Conference on Machine Learning}, pp.\
  3499--3508. PMLR, 2019.

\bibitem[Krenn et~al.(2020)Krenn, H{\"a}se, Nigam, Friederich, and
  Aspuru-Guzik]{krenn2020self}
Krenn, M., H{\"a}se, F., Nigam, A., Friederich, P., and Aspuru-Guzik, A.
\newblock Self-referencing embedded strings (selfies): A 100\% robust molecular
  string representation.
\newblock \emph{Machine Learning: Science and Technology}, 1\penalty0
  (4):\penalty0 045024, 2020.

\bibitem[Langevin et~al.(2020)Langevin, Minoux, Levesque, and
  Bianciotto]{doi:10.1021/acs.jcim.0c01015}
Langevin, M., Minoux, H., Levesque, M., and Bianciotto, M.
\newblock Scaffold-constrained molecular generation.
\newblock \emph{Journal of Chemical Information and Modeling}, 60\penalty0
  (12):\penalty0 5637--5646, 2020.
\newblock \doi{10.1021/acs.jcim.0c01015}.
\newblock PMID: 33301333.

\bibitem[Li et~al.(2018)Li, Zhang, and Liu]{li2018multi}
Li, Y., Zhang, L., and Liu, Z.
\newblock Multi-objective de novo drug design with conditional graph generative
  model.
\newblock \emph{Journal of cheminformatics}, 10:\penalty0 1--24, 2018.

\bibitem[Mahmood et~al.(2021)Mahmood, Mansimov, Bonneau, and
  Cho]{mahmood2021masked}
Mahmood, O., Mansimov, E., Bonneau, R., and Cho, K.
\newblock Masked graph modeling for molecule generation.
\newblock \emph{Nature communications}, 12\penalty0 (1):\penalty0 3156, 2021.

\bibitem[Maziarka et~al.(2019)Maziarka, Danel, Mucha, Rataj, Tabor, and
  Jastrzebski]{maziarka2019molecule}
Maziarka, {\L}., Danel, T., Mucha, S., Rataj, K., Tabor, J., and Jastrzebski,
  S.
\newblock Molecule-augmented attention transformer.
\newblock In \emph{Workshop on Graph Representation Learning, Neural
  Information Processing Systems}, 2019.

\bibitem[Mazuz et~al.(2023)Mazuz, Shtar, Shapira, and
  Rokach]{mazuz2023molecule}
Mazuz, E., Shtar, G., Shapira, B., and Rokach, L.
\newblock Molecule generation using transformers and policy gradient
  reinforcement learning.
\newblock \emph{Scientific Reports}, 13\penalty0 (1):\penalty0 8799, 2023.

\bibitem[Medina et~al.(2023)Medina, Linke, Stoll, and
  Sundmacher]{SanchezMedina2023}
Medina, E. I.~S., Linke, S., Stoll, M., and Sundmacher, K.
\newblock Gibbs{\textendash}helmholtz graph neural network: capturing the
  temperature dependency of activity coefficients at infinite dilution.
\newblock \emph{Digital Discovery}, 2\penalty0 (3):\penalty0 781--798, 2023.
\newblock \doi{10.1039/d2dd00142j}.
\newblock URL \url{https://doi.org/10.1039/d2dd00142j}.

\bibitem[Mnih(2013)]{mnih2013playing}
Mnih, V.
\newblock Playing atari with deep reinforcement learning.
\newblock \emph{arXiv preprint arXiv:1312.5602}, 2013.

\bibitem[Nevolianis et~al.(2024)Nevolianis, Rittig, Mitsos, and
  Leonhard]{nevolianis2024multi}
Nevolianis, T., Rittig, J.~G., Mitsos, A., and Leonhard, K.
\newblock Multi-fidelity graph neural networks for predicting toluene/water
  partition coefficients.
\newblock \emph{ChemRxiv preprint 10.26434/chemrxiv-2024-3t818}, 2024.

\bibitem[Nigam et~al.(2021)Nigam, Pollice, Krenn, dos Passos~Gomes, and
  Aspuru-Guzik]{nigam2021beyond}
Nigam, A., Pollice, R., Krenn, M., dos Passos~Gomes, G., and Aspuru-Guzik, A.
\newblock Beyond generative models: superfast traversal, optimization, novelty,
  exploration and discovery (stoned) algorithm for molecules using selfies.
\newblock \emph{Chemical science}, 12\penalty0 (20):\penalty0 7079--7090, 2021.

\bibitem[Nigam et~al.(2023)Nigam, Pollice, Tom, Jorner, Willes, Thiede,
  Kundaje, and Aspuru-Guzik]{nigam2023tartarus}
Nigam, A., Pollice, R., Tom, G., Jorner, K., Willes, J., Thiede, L., Kundaje,
  A., and Aspuru-Guzik, A.
\newblock Tartarus: A benchmarking platform for realistic and practical inverse
  molecular design.
\newblock \emph{Advances in Neural Information Processing Systems},
  36:\penalty0 3263--3306, 2023.

\bibitem[Nnadili et~al.(2024)Nnadili, Okafor, Olayiwola, Akinpelu, Kumar, and
  Romagnoli]{nnadili2024surfactant}
Nnadili, M., Okafor, A.~N., Olayiwola, T., Akinpelu, D., Kumar, R., and
  Romagnoli, J.~A.
\newblock Surfactant-specific ai-driven molecular design: Integrating
  generative models, predictive modeling, and reinforcement learning for
  tailored surfactant synthesis.
\newblock \emph{Industrial \& Engineering Chemistry Research}, 63\penalty0
  (14):\penalty0 6313--6324, 2024.

\bibitem[O'Boyle \& Dalke(2018)O'Boyle and Dalke]{o2018deepsmiles}
O'Boyle, N. and Dalke, A.
\newblock Deepsmiles: an adaptation of smiles for use in machine-learning of
  chemical structures.
\newblock 2018.

\bibitem[Olivecrona et~al.(2017)Olivecrona, Blaschke, Engkvist, and
  Chen]{olivecrona2017molecular}
Olivecrona, M., Blaschke, T., Engkvist, O., and Chen, H.
\newblock Molecular de-novo design through deep reinforcement learning.
\newblock \emph{Journal of cheminformatics}, 9:\penalty0 1--14, 2017.

\bibitem[Peters et~al.(2008)Peters, Zavrel, Kahlen, Schmidt,
  Ansorge-Schumacher, Leitner, B{\"u}chs, Greiner, and Spiess]{Peters.2008}
Peters, M., Zavrel, M., Kahlen, J., Schmidt, T., Ansorge-Schumacher, M.,
  Leitner, W., B{\"u}chs, J., Greiner, L., and Spiess, A.~C.
\newblock Systematic approach to solvent selection for biphasic systems with a
  combination of cosmo--rs and a dynamic modeling tool.
\newblock \emph{Engineering in Life Sciences}, 8\penalty0 (5):\penalty0
  546--552, 2008.
\newblock ISSN 1618-0240.
\newblock \doi{10.1002/elsc.200800037}.

\bibitem[Pirnay \& Grimm(2024{\natexlab{a}})Pirnay and
  Grimm]{pirnay2024selfimprovement}
Pirnay, J. and Grimm, D.~G.
\newblock Self-improvement for neural combinatorial optimization: Sample
  without replacement, but improvement.
\newblock \emph{Transactions on Machine Learning Research}, 2024{\natexlab{a}}.
\newblock ISSN 2835-8856.
\newblock URL \url{https://openreview.net/forum?id=agT8ojoH0X}.
\newblock Featured Certification.

\bibitem[Pirnay \& Grimm(2024{\natexlab{b}})Pirnay and Grimm]{pirnay2024take}
Pirnay, J. and Grimm, D.~G.
\newblock Take a step and reconsider: Sequence decoding for self-improved
  neural combinatorial optimization.
\newblock In \emph{European Conference on Artificial Intelligence 2024}, pp.\
  1927--1934. ECAI, 2024{\natexlab{b}}.

\bibitem[Polykovskiy et~al.(2020)Polykovskiy, Zhebrak, Sanchez-Lengeling,
  Golovanov, Tatanov, Belyaev, Kurbanov, Artamonov, Aladinskiy, Veselov,
  et~al.]{polykovskiy2020molecular}
Polykovskiy, D., Zhebrak, A., Sanchez-Lengeling, B., Golovanov, S., Tatanov,
  O., Belyaev, S., Kurbanov, R., Artamonov, A., Aladinskiy, V., Veselov, M.,
  et~al.
\newblock Molecular sets (moses): a benchmarking platform for molecular
  generation models.
\newblock \emph{Frontiers in Pharmacology}, 11:\penalty0 565644, 2020.

\bibitem[Ramakrishnan et~al.(2014)Ramakrishnan, Dral, Rupp, and von
  Lilienfeld]{ramakrishnan2014quantum}
Ramakrishnan, R., Dral, P.~O., Rupp, M., and von Lilienfeld, O.~A.
\newblock Quantum chemistry structures and properties of 134 kilo molecules.
\newblock \emph{Scientific Data}, 1, 2014.

\bibitem[Reiser et~al.(2022)Reiser, Neubert, Eberhard, Torresi, Zhou, Shao,
  Metni, {van Hoesel}, Schopmans, Sommer, and Friederich]{Reiser.2022}
Reiser, P., Neubert, M., Eberhard, A., Torresi, L., Zhou, C., Shao, C., Metni,
  H., {van Hoesel}, C., Schopmans, H., Sommer, T., and Friederich, P.
\newblock Graph neural networks for materials science and chemistry.
\newblock \emph{Communications Materials}, 3\penalty0 (1):\penalty0 93, 2022.
\newblock \doi{10.1038/s43246-022-00315-6}.

\bibitem[Rittig \& Mitsos(2024)Rittig and Mitsos]{Rittig.2024}
Rittig, J.~G. and Mitsos, A.
\newblock Thermodynamics-consistent graph neural networks.
\newblock \emph{Chem. Sci.}, 15:\penalty0 18504--18512, 2024.
\newblock \doi{10.1039/D4SC04554H}.
\newblock URL \url{http://dx.doi.org/10.1039/D4SC04554H}.

\bibitem[Rittig et~al.()Rittig, Franke, and Mitsos]{rittig2024deterministic}
Rittig, J.~G., Franke, M., and Mitsos, A.
\newblock Deterministic global optimization for sample-efficient molecular
  design with generative machine learning.
\newblock In \emph{AI for Accelerated Materials Design-NeurIPS 2024}.

\bibitem[Rittig et~al.(2023{\natexlab{a}})Rittig, {Ben Hicham}, Schweidtmann,
  Dahmen, and Mitsos]{Rittig2023a}
Rittig, J.~G., {Ben Hicham}, K., Schweidtmann, A.~M., Dahmen, M., and Mitsos,
  A.
\newblock Graph neural networks for temperature-dependent activity coefficient
  prediction of solutes in ionic liquids.
\newblock \emph{Computers and Chemical Engineering}, 171:\penalty0 108153,
  2023{\natexlab{a}}.
\newblock ISSN 00981354.
\newblock \doi{10.1016/j.compchemeng.2023.108153}.
\newblock URL \url{https://doi.org/10.1016/j.compchemeng.2023.108153}.

\bibitem[Rittig et~al.(2023{\natexlab{b}})Rittig, Felton, Lapkin, and
  Mitsos]{Rittig2023b}
Rittig, J.~G., Felton, K.~C., Lapkin, A.~A., and Mitsos, A.
\newblock {G}ibbs-{D}uhem-informed neural networks for binary activity
  coefficient prediction.
\newblock \emph{Digital Discovery}, 2:\penalty0 1752--1767, 2023{\natexlab{b}}.
\newblock \doi{10.1039/D3DD00103B}.
\newblock URL \url{https://doi.org/10.1039/D3DD00103B}.

\bibitem[Rittig et~al.(2023{\natexlab{c}})Rittig, Gao, Dahmen, Mitsos, and
  Schweidtmann]{Rittig_GNNBook.2022}
Rittig, J.~G., Gao, Q., Dahmen, M., Mitsos, A., and Schweidtmann, A.~M.
\newblock Graph neural networks for the prediction of molecular
  structure--property relationships.
\newblock In Zhang, D. and {Del R{\'i}o Chanona}, E.~A. (eds.), \emph{Machine
  Learning and Hybrid Modelling for Reaction Engineering}, pp.\  159--181.
  {Royal Society of Chemistry}, 2023{\natexlab{c}}.
\newblock ISBN 978-1-83916-563-4.
\newblock \doi{10.1039/BK9781837670178-00159}.

\bibitem[Rittig et~al.(2023{\natexlab{d}})Rittig, Ritzert, Schweidtmann,
  Winkler, Weber, Morsch, Heufer, Grohe, Mitsos, and Dahmen]{rittig2023graph}
Rittig, J.~G., Ritzert, M., Schweidtmann, A.~M., Winkler, S., Weber, J.~M.,
  Morsch, P., Heufer, K.~A., Grohe, M., Mitsos, A., and Dahmen, M.
\newblock Graph machine learning for design of high-octane fuels.
\newblock \emph{AIChE journal}, 69\penalty0 (4):\penalty0 e17971,
  2023{\natexlab{d}}.

\bibitem[Ruddigkeit et~al.(2012)Ruddigkeit, Van~Deursen, Blum, and
  Reymond]{ruddigkeit2012enumeration}
Ruddigkeit, L., Van~Deursen, R., Blum, L.~C., and Reymond, J.-L.
\newblock Enumeration of 166 billion organic small molecules in the chemical
  universe database gdb-17.
\newblock \emph{Journal of chemical information and modeling}, 52\penalty0
  (11):\penalty0 2864--2875, 2012.

\bibitem[Sanchez-Lengeling \& Aspuru-Guzik(2018)Sanchez-Lengeling and
  Aspuru-Guzik]{sanchez2018inverse}
Sanchez-Lengeling, B. and Aspuru-Guzik, A.
\newblock Inverse molecular design using machine learning: Generative models
  for matter engineering.
\newblock \emph{Science}, 361\penalty0 (6400):\penalty0 360--365, 2018.

\bibitem[Sarathy \& Eraqi(2024)Sarathy and Eraqi]{sarathy2024artificial}
Sarathy, S.~M. and Eraqi, B.~A.
\newblock Artificial intelligence for novel fuel design.
\newblock \emph{Proceedings of the Combustion Institute}, 40\penalty0
  (1-4):\penalty0 105630, 2024.

\bibitem[Schilter et~al.(2023)Schilter, Vaucher, Schwaller, and
  Laino]{schilter2023designing}
Schilter, O., Vaucher, A., Schwaller, P., and Laino, T.
\newblock Designing catalysts with deep generative models and computational
  data. a case study for suzuki cross coupling reactions.
\newblock \emph{Digital discovery}, 2\penalty0 (3):\penalty0 728--735, 2023.

\bibitem[Schneider \& Fechner(2005)Schneider and Fechner]{Schneider_2005}
Schneider, G. and Fechner, U.
\newblock Computer-based de novo design of drug-like molecules.
\newblock \emph{Nature Reviews Drug Discovery}, 4\penalty0 (8):\penalty0
  649--663, 2005.

\bibitem[Schulman et~al.(2017)Schulman, Wolski, Dhariwal, Radford, and
  Klimov]{schulman2017proximal}
Schulman, J., Wolski, F., Dhariwal, P., Radford, A., and Klimov, O.
\newblock Proximal policy optimization algorithms.
\newblock \emph{arXiv preprint arXiv:1707.06347}, 2017.

\bibitem[Segler et~al.(2018)Segler, Kogej, Tyrchan, and
  Waller]{segler2018generating}
Segler, M.~H., Kogej, T., Tyrchan, C., and Waller, M.~P.
\newblock Generating focused molecule libraries for drug discovery with
  recurrent neural networks.
\newblock \emph{ACS central science}, 4\penalty0 (1):\penalty0 120--131, 2018.

\bibitem[Shi et~al.(2020)Shi, Bieber, and Sutton]{shi2020incremental}
Shi, K., Bieber, D., and Sutton, C.
\newblock Incremental sampling without replacement for sequence models.
\newblock In \emph{International Conference on Machine Learning}, pp.\
  8785--8795. PMLR, 2020.

\bibitem[Skinnider(2024)]{skinnider2024invalid}
Skinnider, M.~A.
\newblock Invalid smiles are beneficial rather than detrimental to chemical
  language models.
\newblock \emph{Nature Machine Intelligence}, 6\penalty0 (4):\penalty0
  437--448, 2024.

\bibitem[Specht et~al.(2024)Specht, Nagda, Fellenz, Mandt, Hasse, and
  Jirasek]{specht2024hanna}
Specht, T., Nagda, M., Fellenz, S., Mandt, S., Hasse, H., and Jirasek, F.
\newblock Hanna: Hard-constraint neural network for consistent activity
  coefficient prediction.
\newblock \emph{arXiv preprint arXiv:2407.18011}, 2024.

\bibitem[Thomas et~al.(2022)Thomas, O'Boyle, Bender, and
  De~Graaf]{thomas2022re}
Thomas, M., O'Boyle, N.~M., Bender, A., and De~Graaf, C.
\newblock Re-evaluating sample efficiency in de novo molecule generation.
\newblock \emph{arXiv preprint arXiv:2212.01385}, 2022.

\bibitem[Thomas et~al.(2024)Thomas, O’Boyle, Bender, and
  De~Graaf]{thomas2024molscore}
Thomas, M., O’Boyle, N.~M., Bender, A., and De~Graaf, C.
\newblock Molscore: a scoring, evaluation and benchmarking framework for
  generative models in de novo drug design.
\newblock \emph{Journal of Cheminformatics}, 16\penalty0 (1):\penalty0 64,
  2024.

\bibitem[Vaswani(2017)]{vaswani2017attention}
Vaswani, A.
\newblock Attention is all you need.
\newblock \emph{Advances in Neural Information Processing Systems}, 2017.

\bibitem[Wagner(2021)]{wagner2021constructions}
Wagner, A.~Z.
\newblock Constructions in combinatorics via neural networks.
\newblock \emph{arXiv preprint arXiv:2104.14516}, 2021.

\bibitem[Wang et~al.(2023)Wang, Li, and Barati~Farimani]{Wang_2023}
Wang, Y., Li, Z., and Barati~Farimani, A.
\newblock Graph neural networks for molecules.
\newblock In \emph{Machine Learning in Molecular Sciences}, pp.\  21--66.
  Springer, 2023.

\bibitem[Weininger(1988)]{doi:10.1021/ci00057a005}
Weininger, D.
\newblock Smiles, a chemical language and information system. 1. introduction
  to methodology and encoding rules.
\newblock \emph{Journal of Chemical Information and Computer Sciences},
  28\penalty0 (1):\penalty0 31--36, 1988.
\newblock \doi{10.1021/ci00057a005}.
\newblock URL \url{https://doi.org/10.1021/ci00057a005}.

\bibitem[Williams(1992)]{williams1992simple}
Williams, R.~J.
\newblock Simple statistical gradient-following algorithms for connectionist
  reinforcement learning.
\newblock \emph{Machine learning}, 8:\penalty0 229--256, 1992.

\bibitem[Winter et~al.(2022)Winter, Winter, Schilling, and Bardow]{Winter.2022}
Winter, B., Winter, C., Schilling, J., and Bardow, A.
\newblock A smile is all you need: predicting limiting activity coefficients
  from {SMILES} with natural language processing.
\newblock \emph{Digital Discovery}, 1:\penalty0 859--869, 2022.
\newblock \doi{10.1039/d2dd00058j}.

\bibitem[Winter et~al.(2023)Winter, Winter, Esper, Schilling, and
  Bardow]{Winter.2023}
Winter, B., Winter, C., Esper, T., Schilling, J., and Bardow, A.
\newblock Spt-nrtl: A physics-guided machine learning model to predict
  thermodynamically consistent activity coefficients.
\newblock \emph{Fluid Phase Equilibria}, 568:\penalty0 113731, 2023.
\newblock ISSN 03783812.
\newblock \doi{10.1016/j.fluid.2023.113731}.

\bibitem[Wisniak(1983)]{wisniak1983liquid}
Wisniak, J.
\newblock Liquid—liquid phase splitting—i analytical models for critical
  mixing and azeotropy.
\newblock \emph{Chemical Engineering Science}, 38\penalty0 (6):\penalty0
  969--978, 1983.

\bibitem[Xu et~al.(2024)Xu, Fu, Gao, and Sun]{xu2024reinventtransformer}
Xu, P., Fu, T., Gao, W., and Sun, J.
\newblock {REINVENT}-transformer: Molecular de novo design through
  transformer-based reinforcement learning.
\newblock In \emph{Artificial Intelligence and Data Science for Healthcare:
  Bridging Data-Centric AI and People-Centric Healthcare}, 2024.
\newblock URL \url{https://openreview.net/forum?id=XykiSFid41}.

\bibitem[Ying et~al.(2021)Ying, Cai, Luo, Zheng, Ke, He, Shen, and
  Liu]{ying2021transformers}
Ying, C., Cai, T., Luo, S., Zheng, S., Ke, G., He, D., Shen, Y., and Liu, T.-Y.
\newblock Do transformers really perform badly for graph representation?
\newblock \emph{Advances in neural information processing systems},
  34:\penalty0 28877--28888, 2021.

\bibitem[Yue et~al.(2025)Yue, Tao, Varshney, and Li]{yue2024benchmarking}
Yue, T., Tao, L., Varshney, V., and Li, Y.
\newblock Benchmarking study of deep generative models for inverse polymer
  design.
\newblock \emph{Digital Discovery}, pp.\ ~--, 2025.
\newblock \doi{10.1039/D4DD00395K}.
\newblock URL \url{http://dx.doi.org/10.1039/D4DD00395K}.

\bibitem[Zamora et~al.(2023)Zamora, Viayna, Pinheiro, Curutchet, Bisbal, Ruiz,
  Ràfols, and Luque]{Zamora2023}
Zamora, W.~J., Viayna, A., Pinheiro, S., Curutchet, C., Bisbal, L., Ruiz, R.,
  Ràfols, C., and Luque, F.~J.
\newblock Prediction of toluene/water partition coefficients in the sampl9
  blind challenge: assessment of machine learning and ief-pcm/mst continuum
  solvation models.
\newblock \emph{Physical Chemistry Chemical Physics}, pp.\  10.1039/D3CP01428B,
  2023.
\newblock \doi{10.1039/D3CP01428B}.
\newblock URL \url{http://dx.doi.org/10.1039/D3CP01428B}.

\bibitem[Zang \& Wang(2020)Zang and Wang]{zang2020moflow}
Zang, C. and Wang, F.
\newblock Moflow: an invertible flow model for generating molecular graphs.
\newblock In \emph{Proceedings of the 26th ACM SIGKDD international conference
  on knowledge discovery \& data mining}, pp.\  617--626, 2020.

\bibitem[Zhang et~al.(2023{\natexlab{a}})Zhang, Wang, Helwig, Luo, Fu, Xie,
  Liu, Lin, Xu, Yan, et~al.]{zhang2023artificial}
Zhang, X., Wang, L., Helwig, J., Luo, Y., Fu, C., Xie, Y., Liu, M., Lin, Y.,
  Xu, Z., Yan, K., et~al.
\newblock Artificial intelligence for science in quantum, atomistic, and
  continuum systems.
\newblock \emph{arXiv preprint arXiv:2307.08423}, 2023{\natexlab{a}}.

\bibitem[Zhang et~al.(2023{\natexlab{b}})Zhang, Liu, Lee, Hsieh, and
  Chen]{Zhang_2023}
Zhang, Z., Liu, Q., Lee, C.-K., Hsieh, C.-Y., and Chen, E.
\newblock An equivariant generative framework for molecular graph-structure
  co-design.
\newblock \emph{Chemical Science}, 14\penalty0 (31):\penalty0 8380--8392,
  2023{\natexlab{b}}.

\bibitem[Zhou et~al.(2019)Zhou, Kearnes, Li, Zare, and
  Riley]{zhou2019optimization}
Zhou, Z., Kearnes, S., Li, L., Zare, R.~N., and Riley, P.
\newblock Optimization of molecules via deep reinforcement learning.
\newblock \emph{Scientific reports}, 9\penalty0 (1):\penalty0 10752, 2019.

\end{thebibliography}
\bibliographystyle{icml2023}

\newpage
\appendix
\onecolumn

\section{Atom alphabet}

We list the atom alphabet for the different design tasks as a Python dictionary. We also provide the valence of an atom, as well as its \verb|formal_charge| and \verb|chiral_tag| – for atom specification in  RDKit – if needed. 

\subsection{Drug design tasks}
\begin{verbatim}
    {
        "C": {"atomic_number": 6, "valence": 4},
        "C-": {"atomic_number": 6, "valence": 3, "formal_charge": -1},
        "C+": {"atomic_number": 6, "valence": 5, "formal_charge": 1},
        "C@": {"atomic_number": 6, "valence": 4, "chiral_tag": CHI_TETRAHEDRAL_CW},
        "C@@": {"atomic_number": 6, "valence": 4, "chiral_tag": CHI_TETRAHEDRAL_CCW},
        "N": {"atomic_number": 7, "valence": 3},
        "N-": {"atomic_number": 7, "valence": 2, "formal_charge": -1},
        "N+": {"atomic_number": 7, "valence": 4, "formal_charge": 1},
        "O": {"atomic_number": 8, "valence": 2},
        "O-": {"atomic_number": 8, "valence": 1, "formal_charge": -1},
        "O+": {"atomic_number": 8, "valence": 3, "formal_charge": 1},
        "F": {"atomic_number": 9, "valence": 1},
        "P": {"atomic_number": 15, "valence": 7},
        "P-": {"atomic_number": 15, "valence": 6, "formal_charge": -1},
        "P+": {"atomic_number": 15, "valence": 8, "formal_charge": 1},
        "S": {"atomic_number": 16, "valence": 6},
        "S-": {"atomic_number": 16, "valence": 5, "formal_charge": -1},
        "S+": {"atomic_number": 16, "valence": 7, "formal_charge": 1},
        "S@": {"atomic_number": 16, "valence": 6, "chiral_tag": CHI_TETRAHEDRAL_CW},
        "S@@": {"atomic_number": 16, "valence": 6, "chiral_tag": CHI_TETRAHEDRAL_CCW},
        "Cl": {"atomic_number": 17, "valence": 1},
        "Br": {"atomic_number": 35, "valence": 1},
        "I": {"atomic_number": 53, "valence": 1}
    }
\end{verbatim}

\subsection{Solvent design tasks}
\begin{verbatim}
    {
        "C": {"atomic_number": 6, "valence": 4},
        "N": {"atomic_number": 7, "valence": 3},
        "O": {"atomic_number": 8, "valence": 2}
    }
\end{verbatim}

\newpage
\section{Drug design results}

Here, we present the complete results for the 20 goal-directed tasks of the GuacaMol benchmark. To put our results into even further perspective, we included additional baseline methods not mentioned in the main paper.

\begin{table*}[ht]
    \caption{GuacaMol goal-directed benchmarks}
    \label{tab:guacamol}
    \centering
    \resizebox{\linewidth}{!}{%
    \begin{tabular}{l c c c c c c c c}
        \toprule
        \textbf{Benchmark} & \textbf{SMILES GA}\ddag\ [1] & \textbf{SMILES LSTM}\ddag\ [2] & \textbf{Graph GA}\ddag\ [3] & \textbf{Reinvent}\P\ [4] & \textbf{GEGL}\P\ [5] & \textbf{CReM} [6] & \textbf{MolRL-MGPT} [7] & \textbf{GraphXForm} (ours) \\
        \midrule
        Celecoxib rediscovery & 0.732 & \textbf{1.000} & \textbf{1.000} & \textbf{1.000} & \textbf{1.000} & \textbf{1.000} & \textbf{1.000} & \textbf{1.000} \\
        Troglitazone rediscovery & 0.515 & \textbf{1.000} & \textbf{1.000} & \textbf{1.000} & 0.552 & \textbf{1.000} & \textbf{1.000} & \textbf{1.000} \\
        Thiothixene rediscovery & 0.598 & \textbf{1.000} & \textbf{1.000} & \textbf{1.000} & \textbf{1.000} & \textbf{1.000} & \textbf{1.000} & \textbf{1.000} \\
        Aripiprazole similarity & 0.834 & \textbf{1.000} & \textbf{1.000} & \textbf{1.000} & \textbf{1.000} & \textbf{1.000} & \textbf{1.000} & \textbf{1.000} \\
        Albuterol similarity & 0.907 & \textbf{1.000} & \textbf{1.000} & \textbf{1.000} & \textbf{1.000} & \textbf{1.000} & \textbf{1.000} & \textbf{1.000} \\
        Mestranol similarity & 0.790 & \textbf{1.000} & \textbf{1.000} & \textbf{1.000} & \textbf{1.000} & \textbf{1.000} & \textbf{1.000} & \textbf{1.000} \\
        C11H24 * & 0.829 & 0.993 & 0.971 & 0.999 & \textbf{1.000} & 0.966 & \textbf{1.000} & \textbf{1.000} \\
        C9H10N2O2PF2Cl \dag & 0.889 & 0.879 & 0.982 & 0.877 & \textbf{1.000} & 0.940 & 0.939 & \textbf{1.000} \\
        Median molecules 1 & 0.334 & 0.438 & 0.406 & 0.434 & 0.455 & 0.371 & 0.449 & \textbf{0.459} \\
        Median molecules 2 & 0.380 & 0.422 & 0.432 & 0.395 & \textbf{0.437} & 0.434 & 0.422 & 0.403 \\
        Osimertinib MPO & 0.886 & 0.907 & 0.953 & 0.889 & \textbf{1.000} & 0.995 & 0.977 & 0.972 \\
        Fexofenadine MPO & 0.931 & 0.959 & 0.998 & \textbf{1.000} & \textbf{1.000} & \textbf{1.000} & \textbf{1.000} & 0.999 \\
        Ranolazine MPO & 0.881 & 0.855 & 0.920 & 0.895 & 0.933 & \textbf{0.969} & 0.939 & 0.944 \\
        Perindopril MPO & 0.661 & 0.808 & 0.792 & 0.764 & 0.833 & 0.815 & 0.810 & \textbf{0.835} \\
        Amlodipine MPO & 0.722 & 0.894 & 0.894 & 0.888 & 0.905 & 0.902 & 0.906 & \textbf{0.923} \\
        Sitagliptin MPO & 0.689 & 0.545 & 0.891 & 0.539 & 0.749 & 0.763 & 0.823 & \textbf{0.965} \\
        Zaleplon MPO & 0.413 & 0.669 & 0.754 & 0.590 & 0.763 & 0.770 & \textbf{0.790} & 0.727 \\
        Valsartan SMARTS & 0.552 & 0.978 & 0.990 & 0.095 & \textbf{1.000} & 0.994 & 0.997 & \textbf{1.000} \\
        Deco hop & 0.970 & 0.996 & \textbf{1.000} & 0.994 & \textbf{1.000} & \textbf{1.000} & \textbf{1.000} & \textbf{1.000} \\
        Scaffold hop & 0.885 & 0.998 & \textbf{1.000} & 0.990 & \textbf{1.000} & \textbf{1.000} & \textbf{1.000} & \textbf{1.000} \\
        \midrule
        \textbf{Total score} & 14.398 & 17.341 & 17.983 & 16.349 & 17.627 & 17.919 & 18.052 & \textbf{18.227} \\
        \bottomrule
        \multicolumn{8}{l}{* Top-159, no stereochemistry. \dag\ Top-250, no stereochemistry. \ddag\ Results as reported in [8]. \P\ Results as reported in [7].} \\
    \end{tabular}
    }
\end{table*}

{\small

[1] Naruki Yoshikawa, Kei Terayama, Masato Sumita, Teruki Homma, Kenta Oono, and Koji Tsuda.
Population-based de novo molecule generation, using grammatical evolution. {\em Chemistry Letters},
47(11):1431–1434, 2018.

[2] Marwin HS Segler, Thierry Kogej, Christian Tyrchan, and Mark P Waller. Generating focused molecule libraries for drug discovery with recurrent neural networks. {\em ACS central science}, 4(1):120–131, 2018.

[3] Jan H Jensen. A graph-based genetic algorithm and generative model/monte carlo tree search for the exploration of chemical space. {\em Chemical science}, 10(12):3567–3572, 2019.

[4] Marcus Olivecrona, Thomas Blaschke, Ola Engkvist, and Hongming Chen. Molecular de-novo design through deep reinforcement learning. {\em Journal of Cheminformatics}, 9, 2017.

[5] Sungsoo Ahn, Junsu Kim, Hankook Lee, and Jinwoo Shin. Guiding deep molecular optimization with genetic exploration. {\em Advances in Neural Information Processing Systems}, 33, 12008–12021, 2020. 

[6] Pavel Polishchuk. CReM: chemically reasonable mutations framework for structure generation. {\em Journal of Cheminformatics}, 12, 2020.

[7] Xiuyuan Hu, Guoqing Liu, Yang Zhao, and Hao Zhang. De novo drug design using reinforcement learning with multiple gpt agents. {\em Advances in Neural Information Processing Systems}, 36, 7405-7418, 2023.

[8] Brown, N.; Fiscato, M.; Segler, M. H. S.; Vaucher, A. C. GuacaMol: Benchmarking Models for de Novo Molecular Design. {\em Journal of Chemical Information and Modeling}, 59(3):1096–1108, 2019.
}

\newpage
\section{Solvent design results}
\subsection{GraphXForm}

\begin{figure}[h]
\centering
\includegraphics[width=\linewidth]{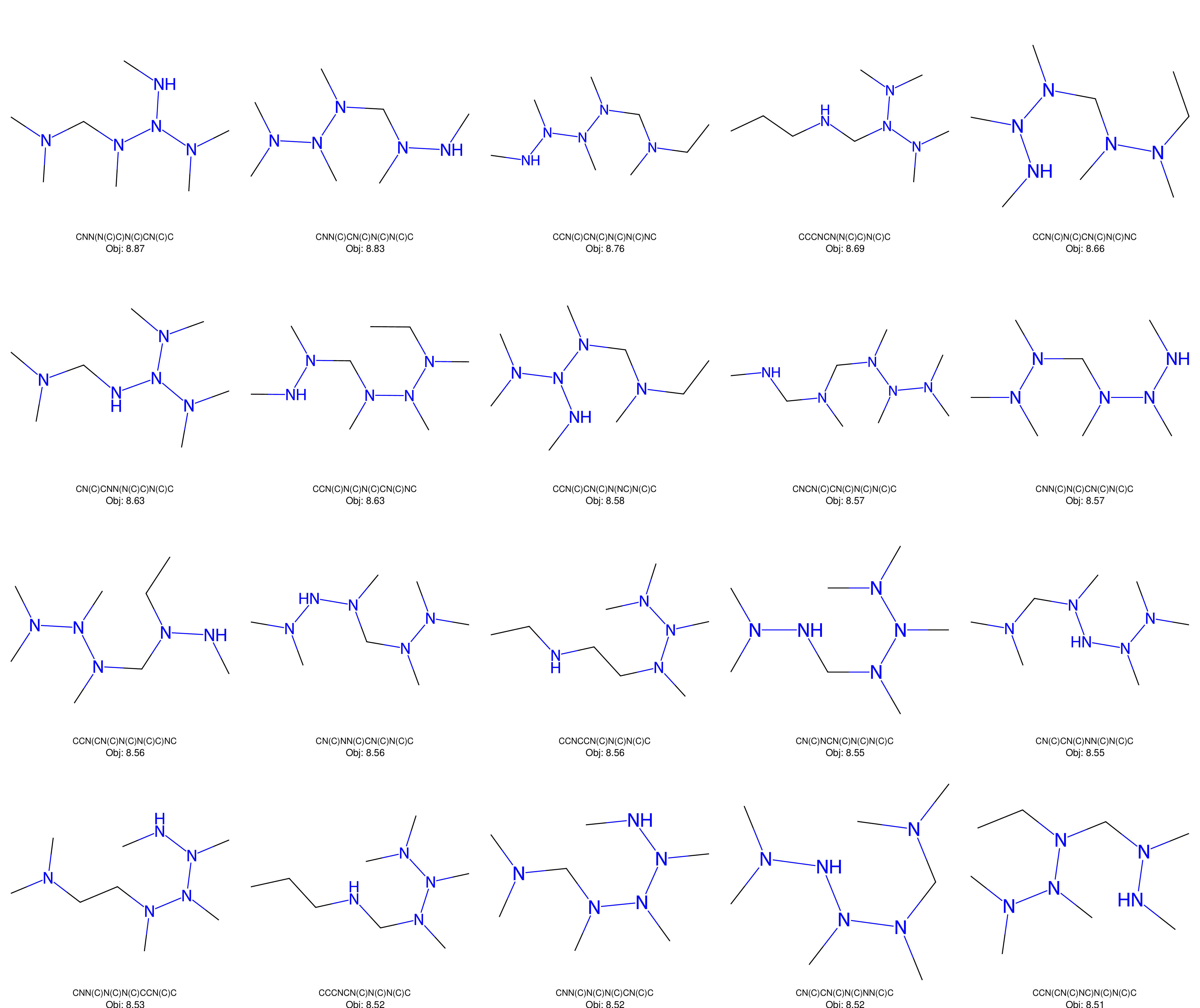}
\caption{\label{fig:ours_iba}Top 20 molecules: GraphXForm, IBA task}
\end{figure}
\begin{figure}
\centering
\includegraphics[width=\linewidth]{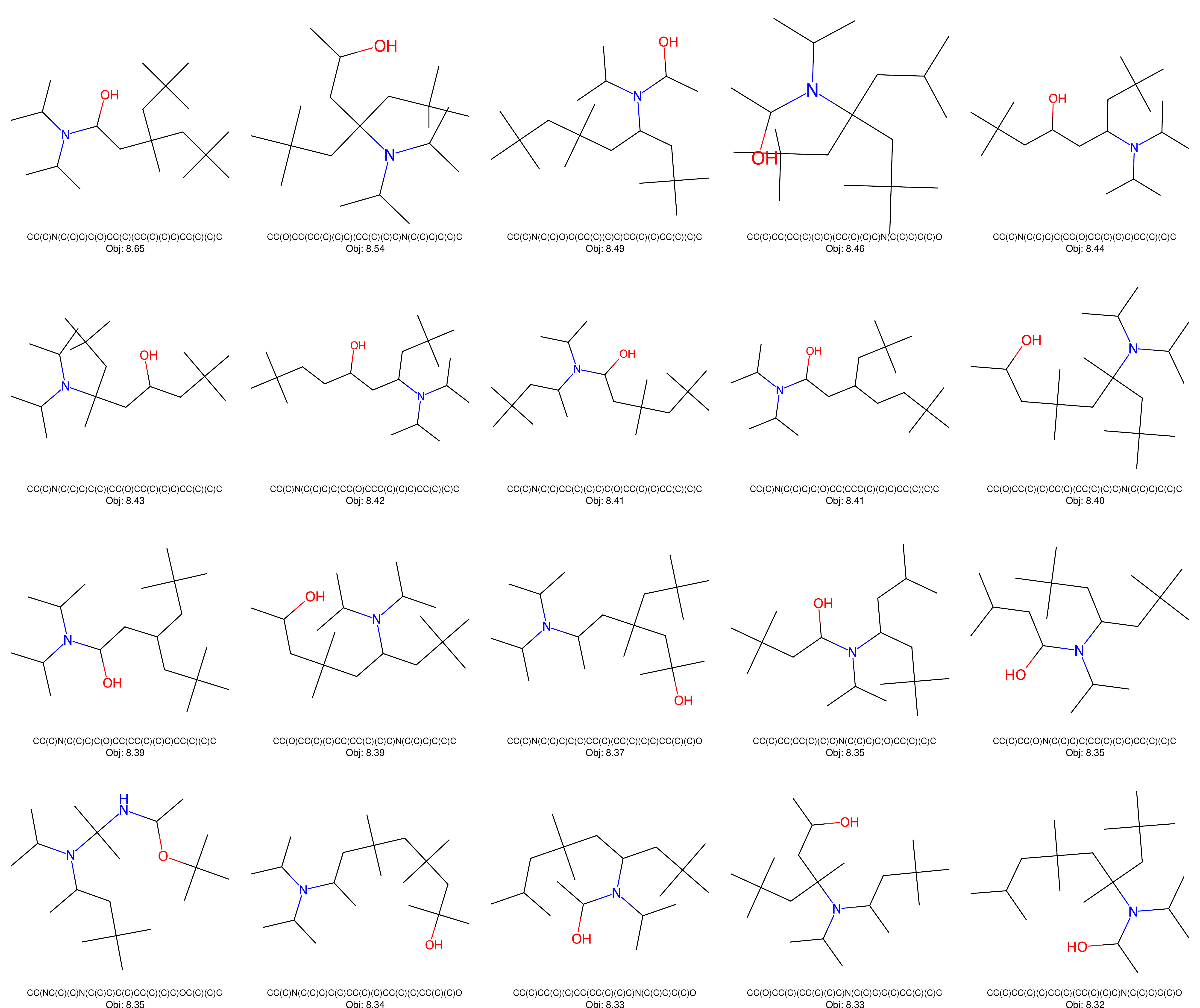}
\caption{\label{fig:ours_tmb}Top 20 molecules: GraphXForm, TMB/DMBA task}
\end{figure}

\newpage
\subsection{List screen}
\begin{figure}[h]
\centering
\includegraphics[width=\linewidth]{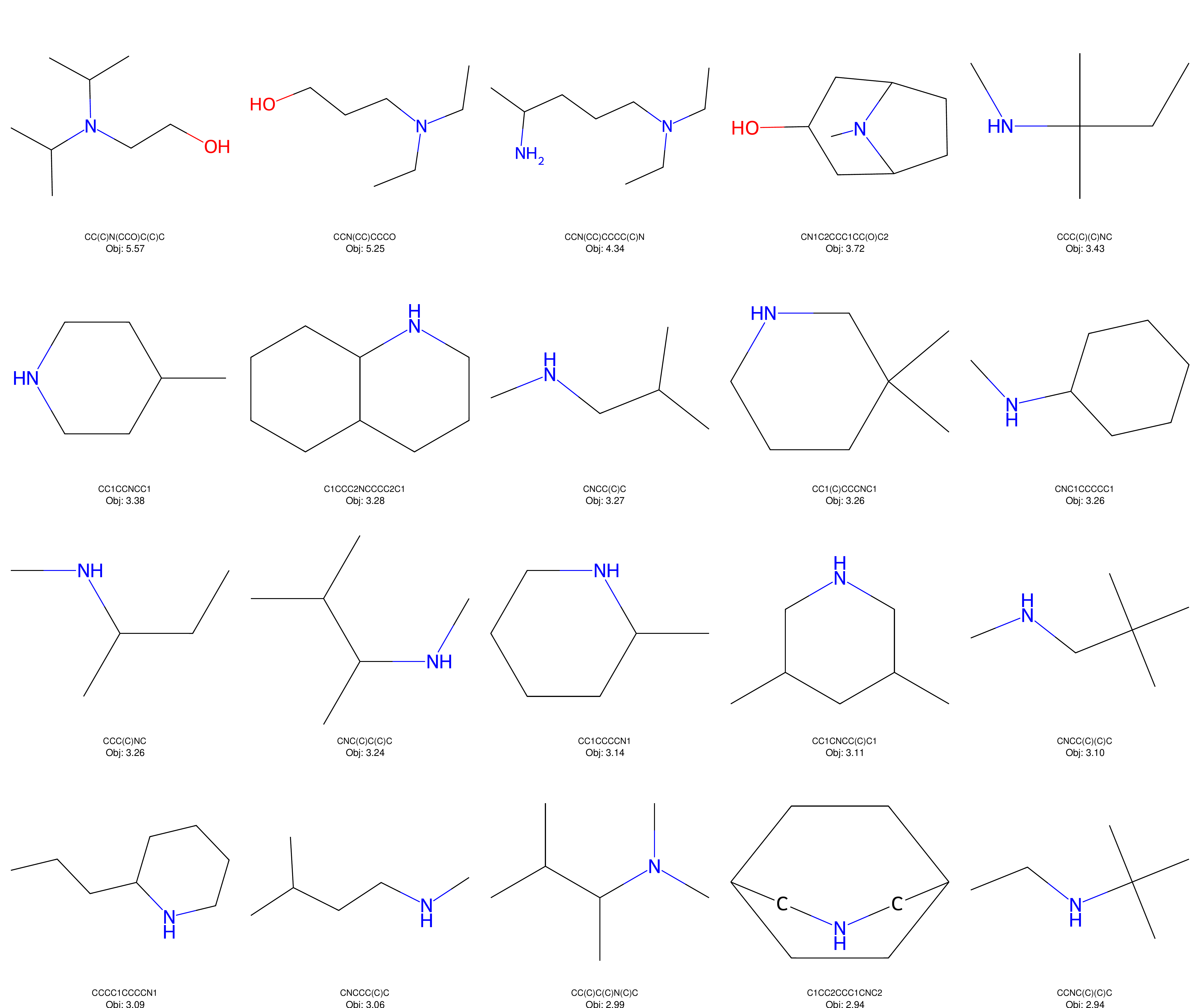}
\caption{\label{fig:list_iba}Top 20 molecules: List screen, IBA task}
\end{figure}
\begin{figure}
\centering
\includegraphics[width=\linewidth]{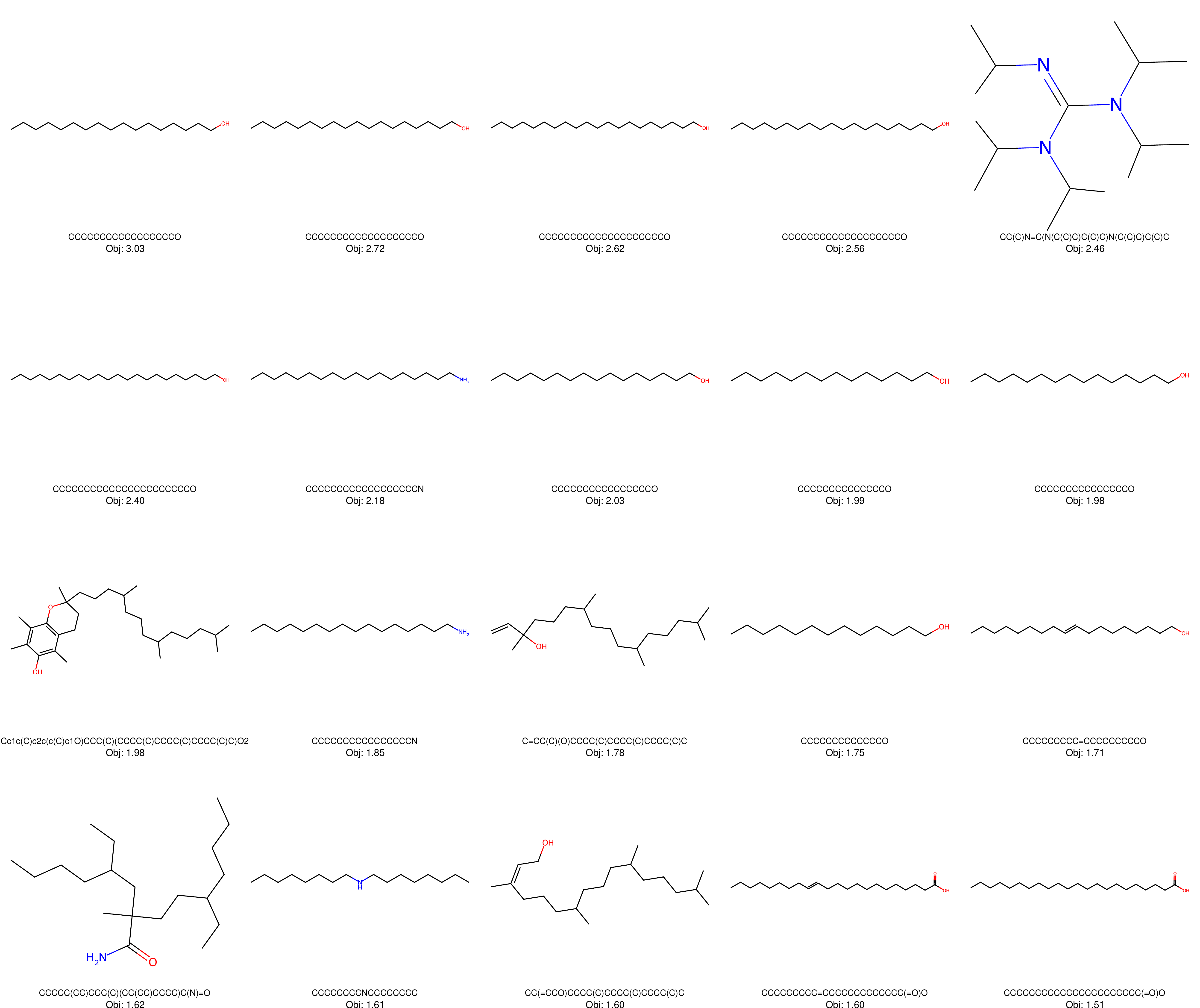}
\caption{\label{fig:list_itmb}Top 20 molecules: List screen, TMB/DMBA task}
\end{figure}

\newpage
\subsection{JT-VAE}
\begin{figure}[h]
\centering
\includegraphics[width=\linewidth]{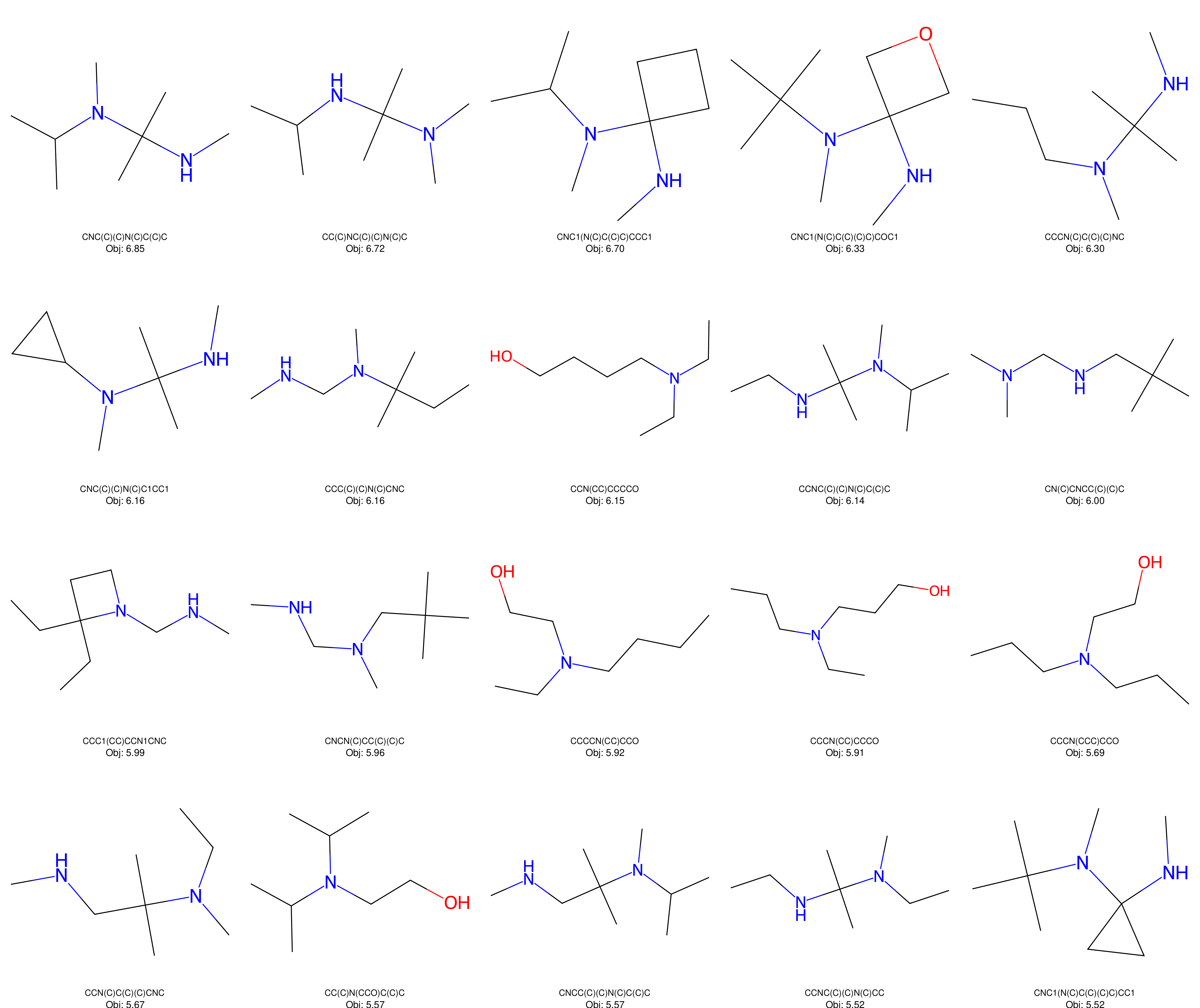}
\caption{\label{fig:jtva_iba}Top 20 molecules: JT-VAE, IBA task}
\end{figure}
\begin{figure}
\centering
\includegraphics[width=\linewidth]{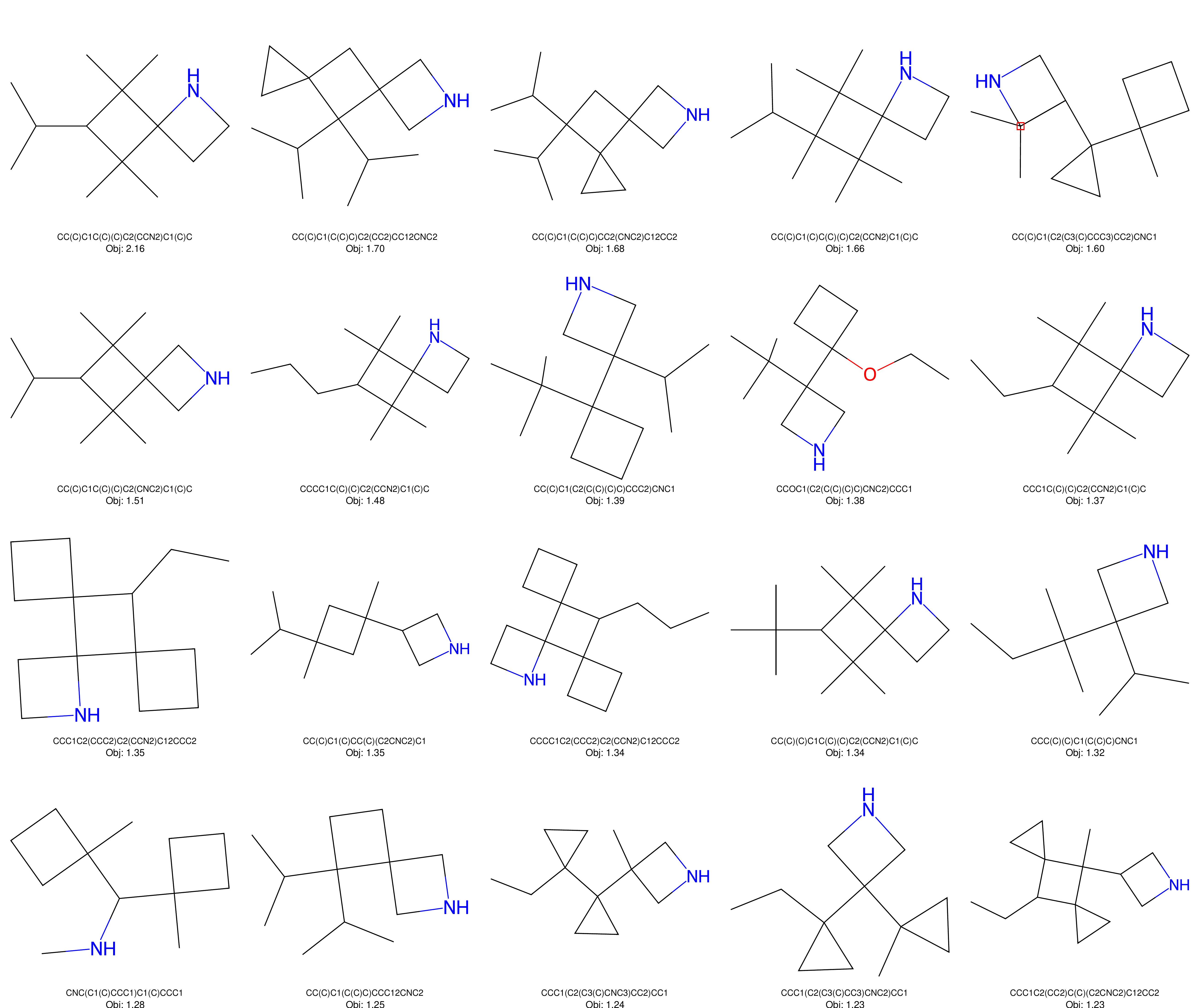}
\caption{\label{fig:jtva_tmb}Top 20 molecules: JT-VAE, TMB/DMBA task}
\end{figure}

\newpage
\subsection{STONED}
\begin{figure}[h]
\centering
\includegraphics[width=\linewidth]{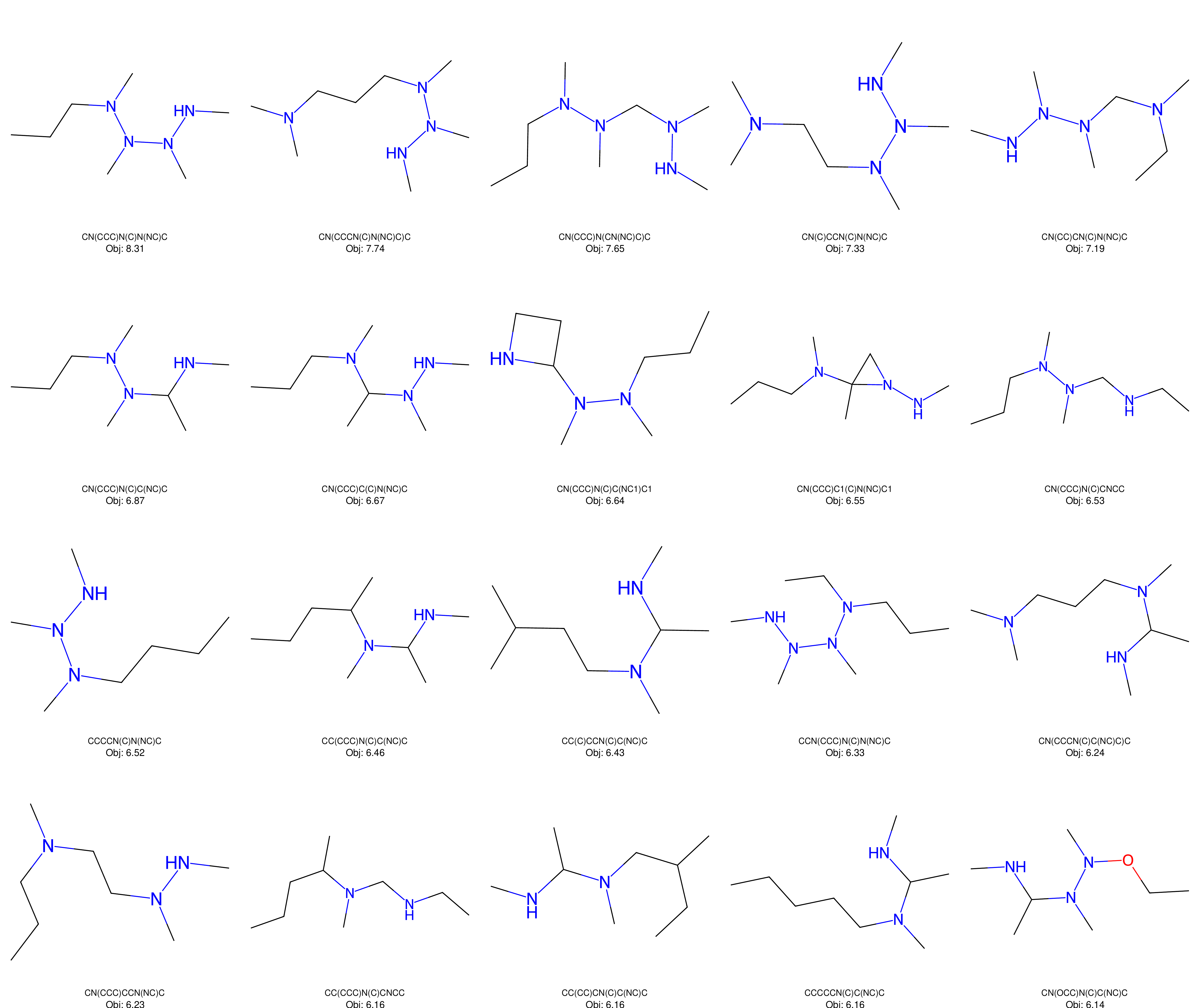}
\caption{\label{fig:ston_iba}Top 20 molecules: STONED, IBA task}
\end{figure}
\begin{figure}
\centering
\includegraphics[width=\linewidth]{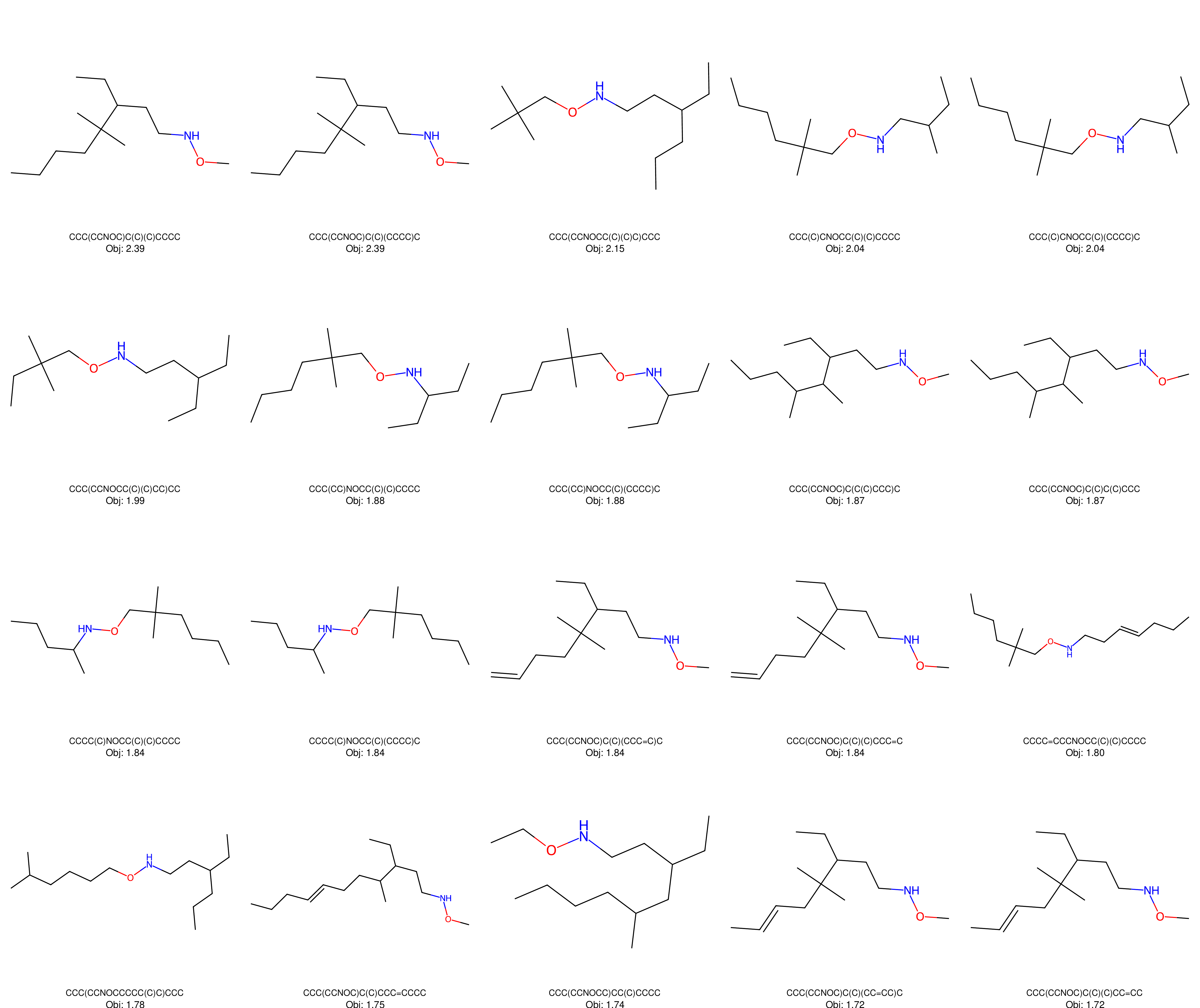}
\caption{\label{fig:ston_tmb}Top 20 molecules: STONED, TMB/DMBA task}
\end{figure}

\newpage
\subsection{Graph GA}
\begin{figure}[h]
\centering
\includegraphics[width=\linewidth]{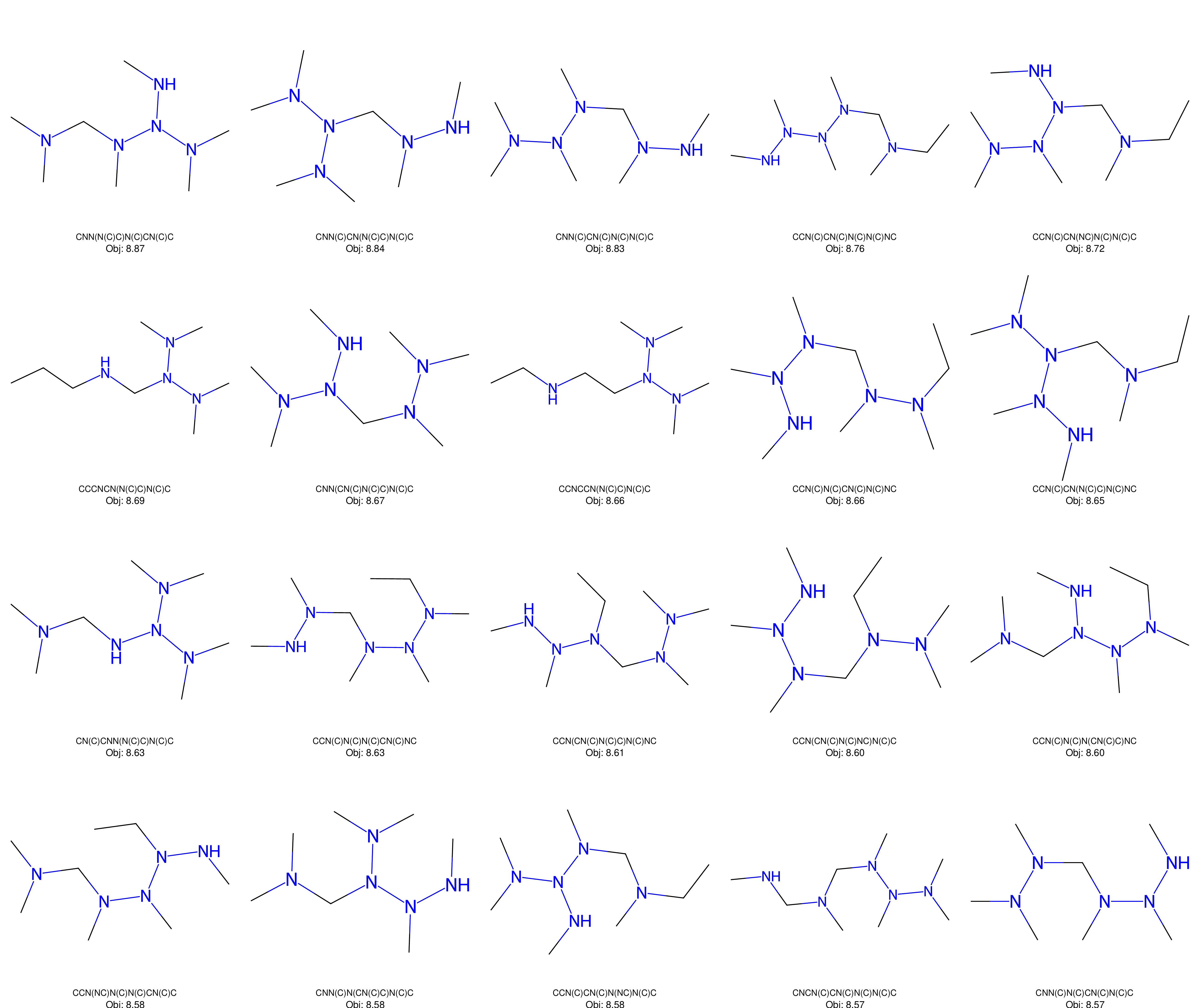}
\caption{\label{fig:grap_iba}Top 20 molecules: Graph GA, IBA task}
\end{figure}
\begin{figure}
\centering
\includegraphics[width=\linewidth]{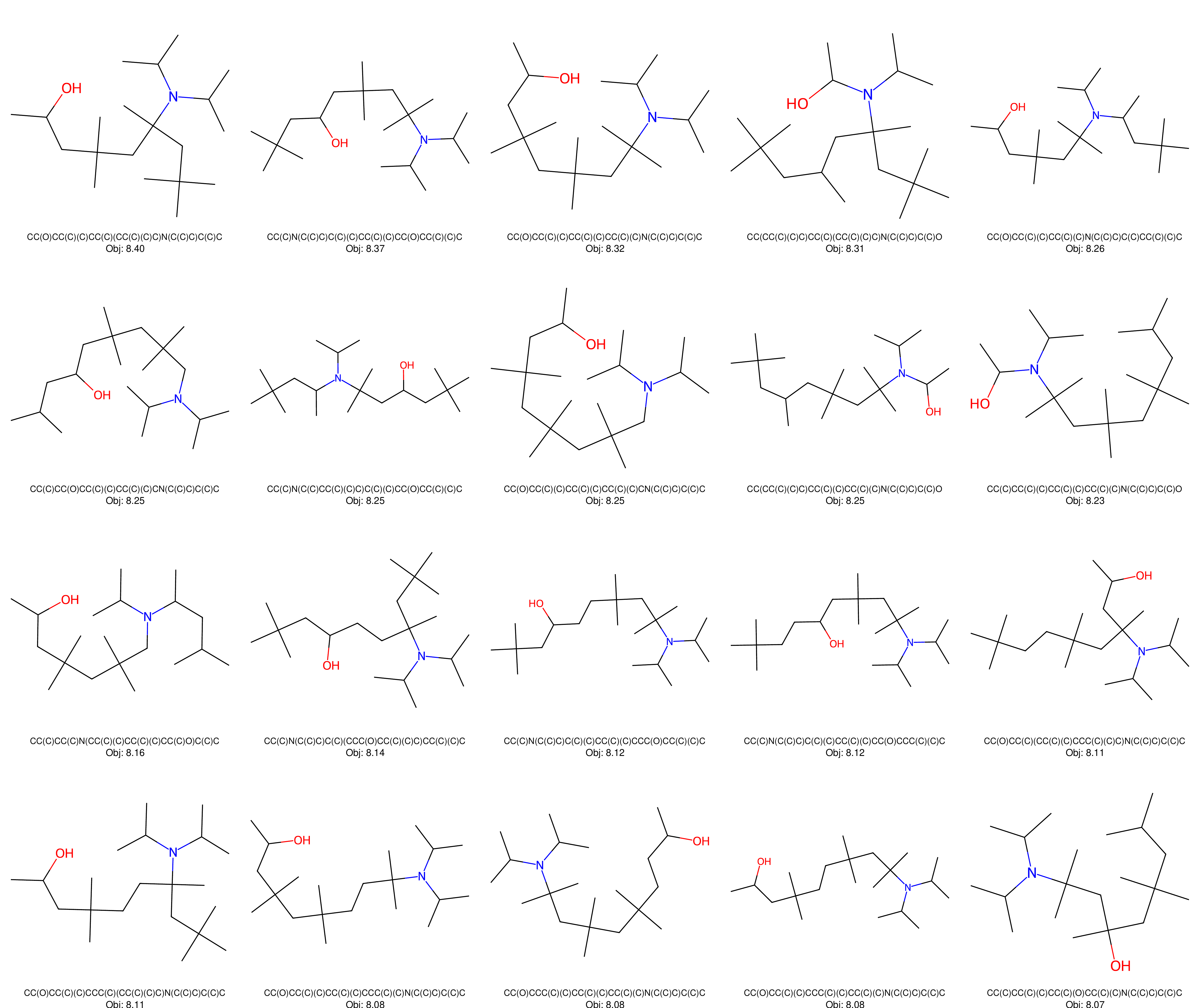}
\caption{\label{fig:grap_tmb}Top 20 molecules: Graph GA, TMB/DMBA task}
\end{figure}

\newpage
\subsection{REINVENT-Transformer}
\begin{figure}[h]
\centering
\includegraphics[width=\linewidth]{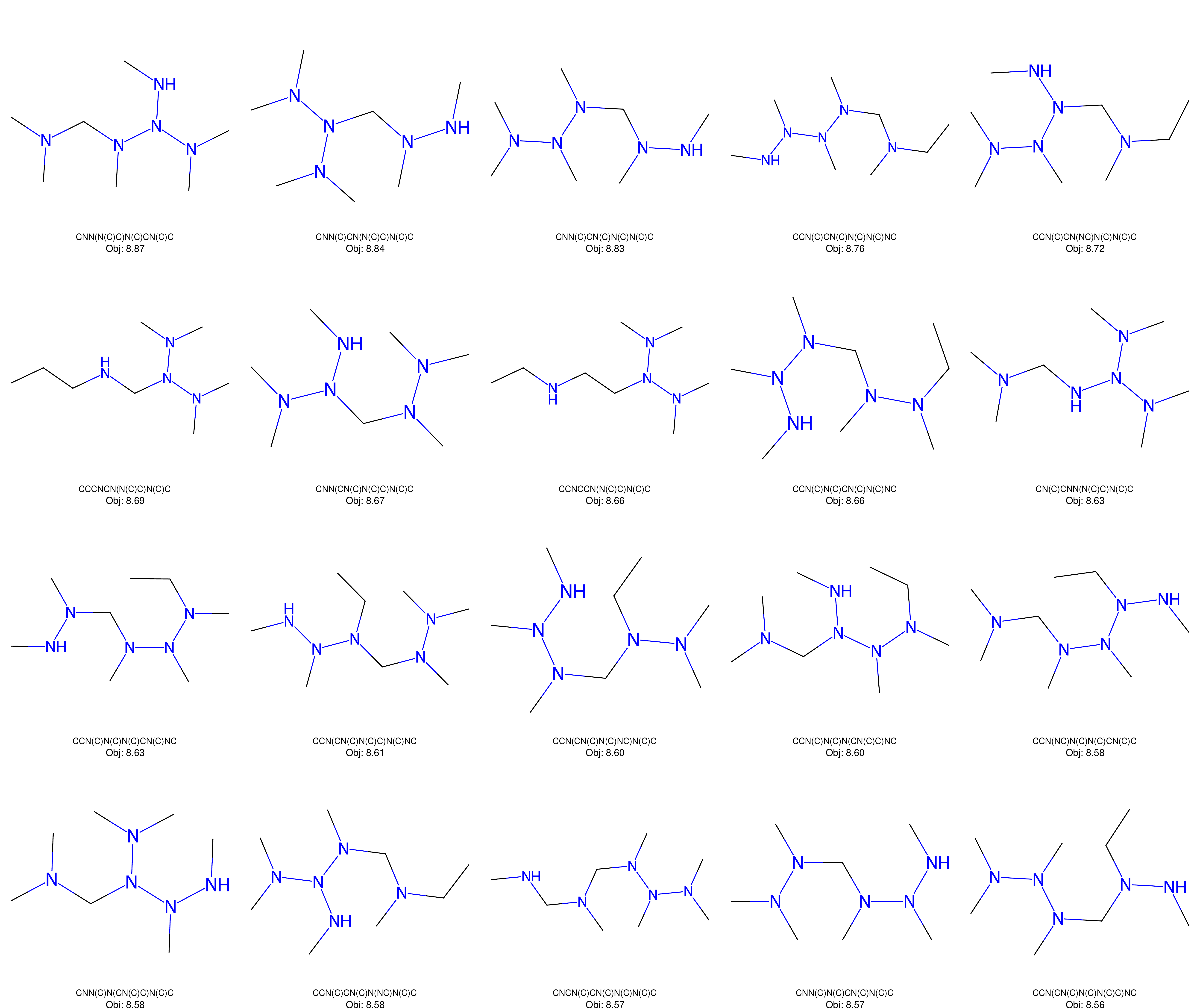}
\caption{\label{fig:rein_iba}Top 20 molecules: REINVENT-Transformer, IBA task}
\end{figure}
\begin{figure}
\centering
\includegraphics[width=\linewidth]{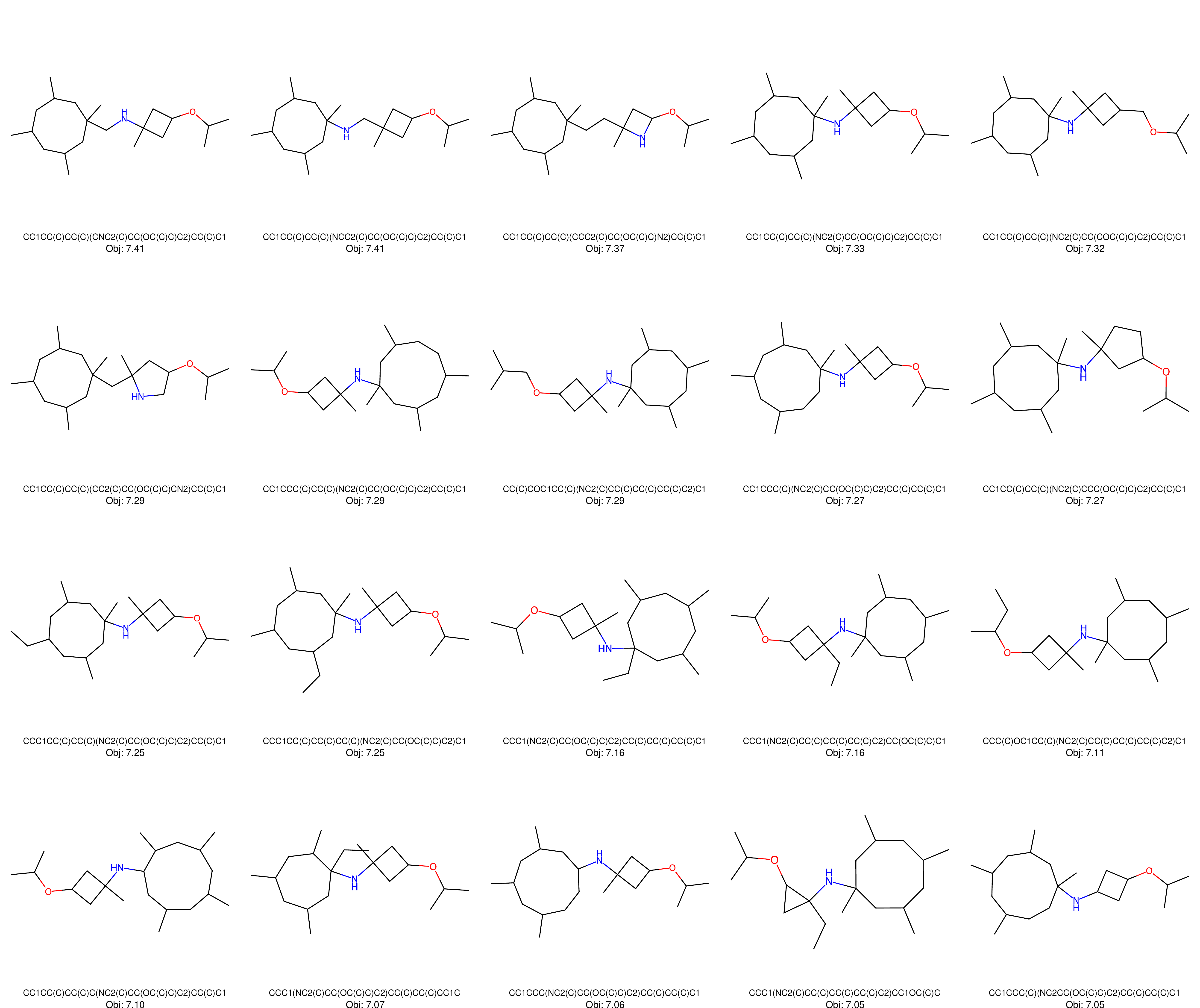}
\caption{\label{fig:rein_tmb}Top 20 molecules: REINVENT-Transformer, TMB/DMBA task}
\end{figure}

\newpage
\subsection{GraphXForm: Structural constraints}
\begin{figure}[h]
\centering
\includegraphics[width=\linewidth]{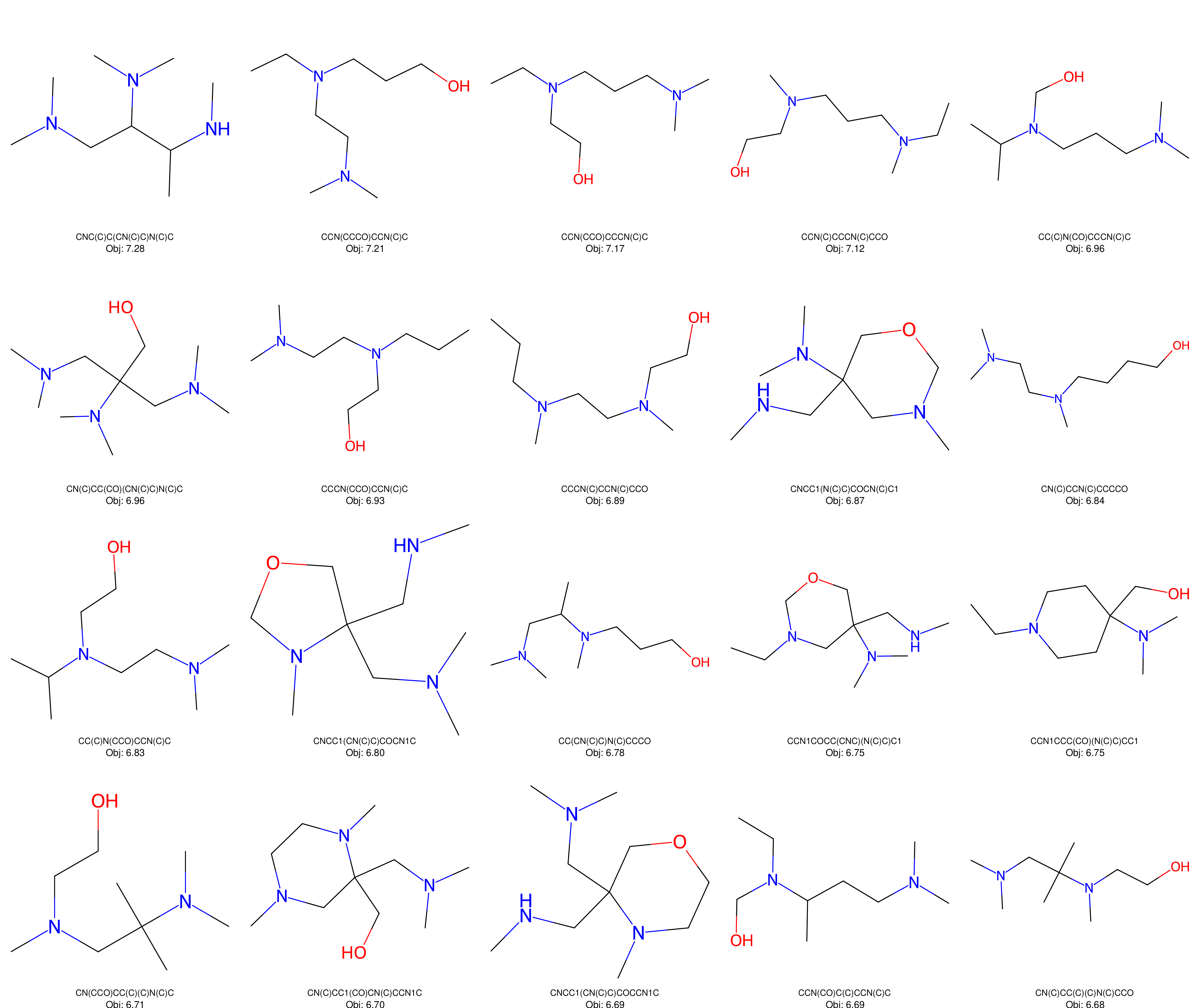}
\caption{\label{fig:stru_iba}Top 20 molecules: GraphXForm, IBA task with structural constraints}
\end{figure}
\begin{figure}
\centering
\includegraphics[width=\linewidth]{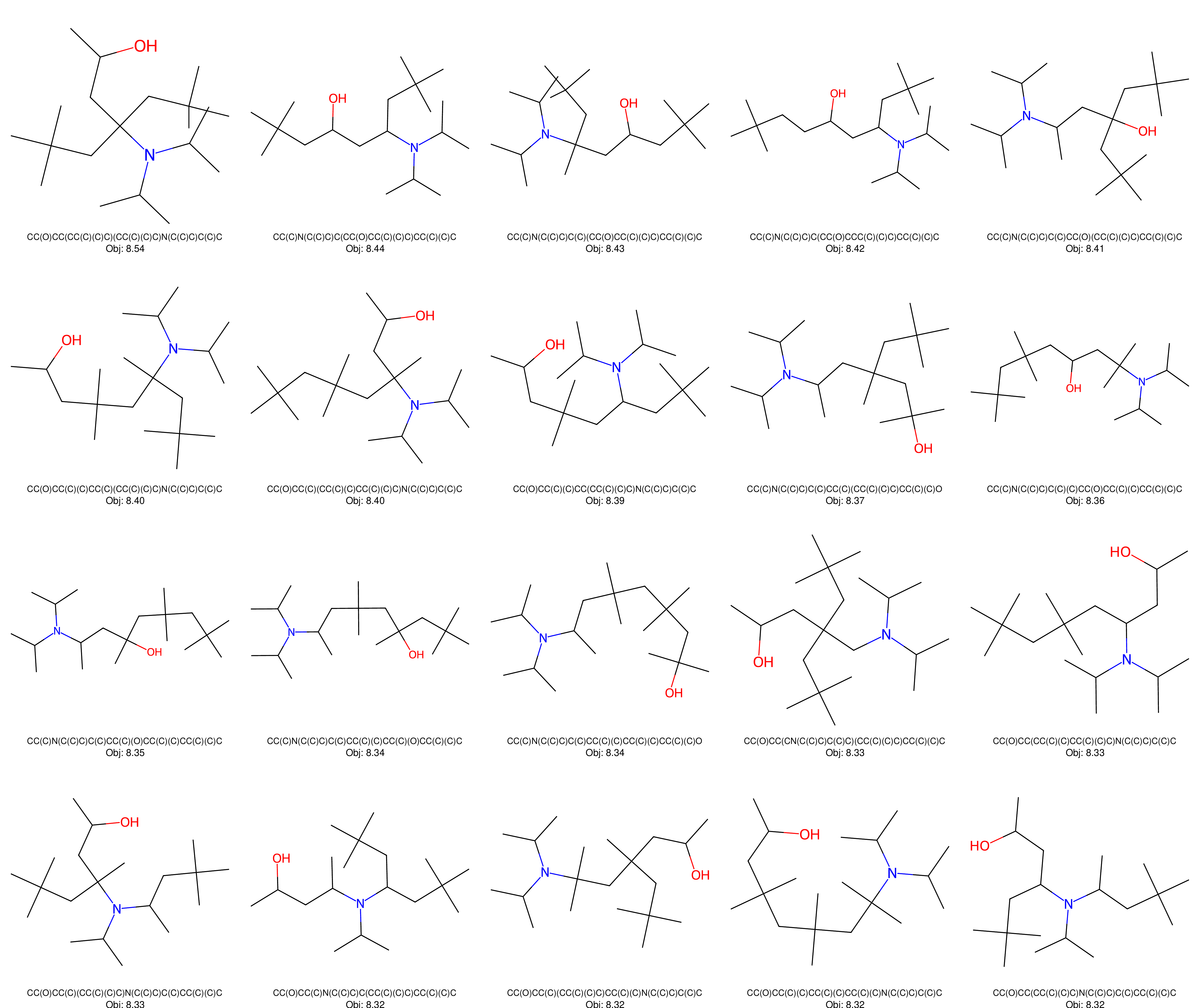}
\caption{\label{fig:stru_tmb}Top 20 molecules: GraphXForm, TMB/DMBA task with structural constraints}
\end{figure}

\end{document}